# A Multi-Resolution Physics-Informed Recurrent Neural Network: Formulation and Application to Musculoskeletal Systems

Karan Taneja[a], Xiaolong He[b], Qizhi He[c] and J. S. Chen[a*]

a Department of Structural Engineering, University of California San Diego, La Jolla, CA, USA

b ANSYS Inc., Livermore, CA, USA

c Department of Civil, Environmental, and Geo- Engineering, University of Minnesota, Minneapolis, MN, USA

## Abstract

This work presents a multi-resolution physics-informed recurrent neural network (MR PI-RNN), for simultaneous prediction of musculoskeletal (MSK) motion and parameter identification of the MSK systems. The MSK application was selected as the model problem due to its challenging nature in mapping the high-frequency surface electromyography (sEMG) signals to the low-frequency body joint motion controlled by the MSK and muscle contraction dynamics. The proposed method utilizes the fast wavelet transform to decompose the mixed frequency input sEMG and output joint motion signals into nested multi-resolution signals. The prediction model is subsequently trained on coarser-scale input-output signals using a gated recurrent unit (GRU), and then the trained parameters are transferred to the next level of training with finer-scale signals. These training processes are repeated recursively under a transfer-learning fashion until the full-scale training (i.e., with unfiltered signals) is achieved, while satisfying the underlying dynamic equilibrium. Numerical examples on recorded subject data demonstrate the effectiveness of the proposed framework in generating a physics-informed forward-dynamics surrogate, which yields higher accuracy in motion predictions of elbow flexion-extension of an MSK system compared to the case with single-scale training. The framework is also capable of identifying muscle parameters that are physiologically consistent with the subject's kinematics data.

---

[*] Corresponding author
*E-mail address*: js-chen@ucsd.edu (J. S. Chen)

## 1. Introduction

The prediction of the evolution of state variables in dynamical systems has been a vital component to several scientific applications such as biology, geophysics, earthquake engineering, solid mechanics, robotics, computer vision [1–7] etc. Black-box techniques based on data-driven mapping and development of parameterized multi-physics models describing the progression of the data have been previously utilized for making predictions on the states. This task continues to be an active area of research due to challenges on many fronts, such as, the quality and scarcity of relevant physical data, the dynamics and complexity of the system, and the reliability and accuracy of the prediction model.

On the other hand, the characterization of parameters in the multi-physics models of these dynamical systems is also critical [8–14]. The task is challenging due in parts to potential noise pollution captured by sensors in the system's measured data, as well as the potential of the parameter space being high-dimensional, leading to ill-posed problems that pose difficulties in numerical solutions. Standard optimization techniques such as genetic algorithms [15,16], simulated annealing [17], and non-linear least squares [18,19] have been employed for parameter identification, but can be computationally expensive and may not converge for ill-posed, non-convex optimization problems that are encountered while solving inverse problems on MSK systems [15,20].

In recent years, machine learning (ML) or deep-learning-based approaches have gained significant popularity for solving forward and inverse problems, attributed to their capability in effectively extracting complex features and patterns from data [21]. This has been successfully demonstrated in numerous engineering applications such as reduced-order modeling [22–26], and materials modeling [27–29], among others. Data-driven computing techniques that enforce constraints of conservation laws in the learning algorithms of a material database, have been developed in the field of computational mechanics [29–37]. More recently, physics-informed neural networks (PINNs) have been developed [11,38,39] to approximate the solutions of given physical equations by using neural networks (NNs). By minimizing the residuals of the governing partial differential equations (PDEs) and the associated initial and boundary conditions, PINNs have been successfully applied to solve forward problems [11,40,41], and inverse problems [11,38,42–44], where the unknown system characteristics are considered trainable parameters or functions [38,45]. For biomechanics and biomedical applications [1,46–50], this method has been applied extensively along with other ML techniques [51,52]. These attempt to bridge the gap between ML-based data-driven surrogate models and the satisfaction of physical laws.

In this study, we focus on the application to musculoskeletal systems, where we aim to utilize non-invasive muscle activity measurements such as surface electromyography (sEMG) signals to predict joint kinetics or kinematics [1,18,19], e.g., for health assessment and rehabilitation purposes [15,16]. These sEMG signals can be used as control inputs to drive the physiological subsystems, which are governed by parameterized non-linear differential equations, that form the forward dynamics problem. Hence, given information on muscle activations, the joint motion of a subject-specific MSK system can be obtained by solving a forward dynamics problem. Data-driven approaches for motion prediction have also been introduced to directly map the input sEMG signal to joint kinetics/kinematics, bypassing the forward dynamics equations and the need for parameter estimation [26–30]. However, the resulting ML-based surrogate models lack interpretability and may not satisfy the underlying physics. Another challenge is that the sEMG signal usually exhibits a wide range of frequencies that are non-trivial for ML models [1] to map to the joint motion.

In our previous work [1], a physics-informed parameter identification neural network (PI-PINN) was proposed for the simultaneous prediction of motion and parameter identification with application to MSK systems. Using the raw transient sEMG signals obtained from the sensors and the corresponding joint motion data, the PI-PINN learned to predict the motion and identify the parameters of the hill-type muscle models representing the contractile muscle-tendon complex. A feature-encoded approach was introduced to enhance the training of the PI-PINN, which yielded high motion prediction accuracy and identified system parameters within a physiological range, with only a limited number of training samples. However, this method relies on mapping in a feature domain constituted by Fourier and polynomial bases, which requires the input sEMG signal to span over the entire duration of the motion. Thus, it prevents *real-time* predictions as the signal is obtained from the sensor.

To enhance the predictive accuracy of the time-dependent signals, recurrent neural networks (RNNs) such as gated recurrent units (GRUs) [29,53,54] are utilized in this study to inform predictions with the history information of the motion. To overcome the limitation of the size of the data and provide more information from the composite frequency bands in the signals, a multi-resolution based (MR) approach is proposed. Wavelets are used to decompose the raw sEMG and joint motion signals into coarse-scale components at various frequency scales and the remaining fine-scale details. Using principles of multi-resolution theory and transfer learning, multi-resolution training processes are repeated recursively from coarse-scale to the full-scale to map the sEMG to the joint motion. To enhance the robustness and generalizability of the model, gaussian noise is introduced to the recorded motion data used for training [29]. The trained model can be applied for real-time motion predictions given the raw sEMG signal obtained from the sensor.

This manuscript is organized as follows. Section 2 introduces the subsystems and mathematical formulations of MSK forward dynamics, followed by an introduction of the proposed multi-resolution PI-RNN framework for simultaneous motion prediction and system parameter identification in Section 3. The following sections verify the proposed framework using synthetic data and validate it by modeling the elbow flexion-extension movement using subject-specific sEMG signals and recorded motion data in Section 4 and 5, respectively. Concluding remarks and future work are summarized in Section 6.

## 2. Formulations for Muscle Mechanics and Musculoskeletal Forward Dynamics

This section provides a brief overview of muscle mechanics and forward dynamics of the human MSK system, with details in Appendix A and B. As depicted in Fig. 1, multiple subsystems within the MSK forward dynamics interact hierarchically: 1) the neural excitation $u(t)$ transforms into muscle activation $a(t)$ (activation dynamics); 2) Muscle activation drives muscle fibers to produce force $F^{MT}$ (muscle-tendon (MT) contraction dynamics); 3) the resultant forces produce joint motion $\boldsymbol{q}$ (translation and rotation) of MSK systems, called the MSK forward dynamics [11,12,37].

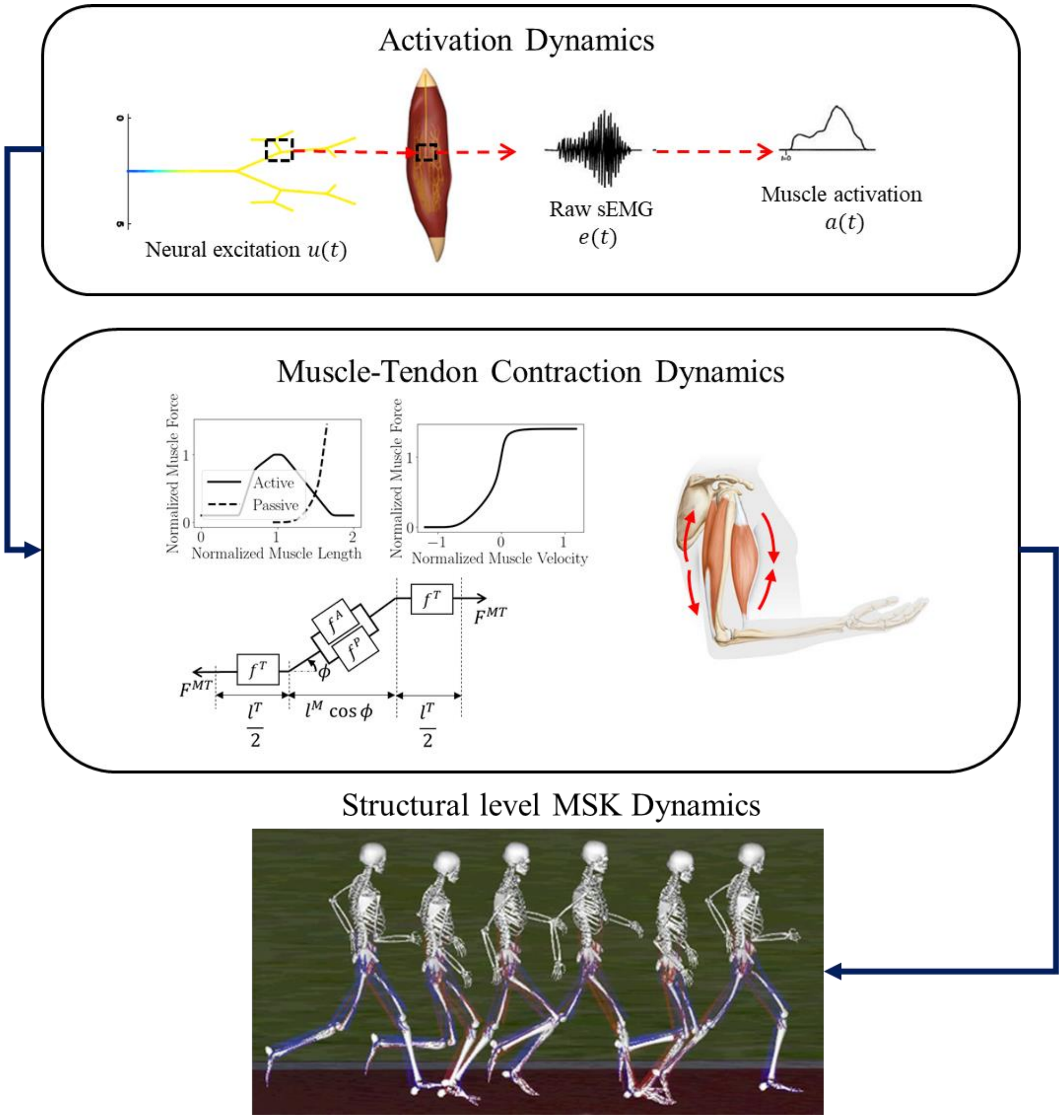

**Fig. 1**: The subsystems involved in the forward dynamics of an MSK system are depicted in this flowchart. Neural excitations are transmitted to muscle fibers (activation dynamics) that contract to produce force (muscle-tendon contraction dynamics). These forces generate torques at the joints (structural level MSK dynamics) leading to joint motion [1,55].

## 2.1 Neural Excitation-to-Activation Dynamics

While activations $a(t)$ in the muscle fibers can be obtained through a non-linear transformation on neural excitations $u(t)$, they are difficult to measure *in-vivo*. Therefore, the excitations are estimated from [15,16] the raw sEMG signals $e(t)$ considering an electro-mechanical delay:

$$u(t) = e(t - d). \tag{1}$$

where $d$ measures the delay between the neural excitation originating and reaching the muscle group. The muscle activation signal $a(t)$ is then expressed as,

$$a(t) = \frac{\exp\left(Au(t)\right) - 1}{\exp(A) - 1} \tag{2}$$

where $A$ is a shape factor. These activations initiate muscle fiber contraction leading to force production from the muscle group.

### 2.2 Muscle-Tendon Force Generation through Contraction Dynamics

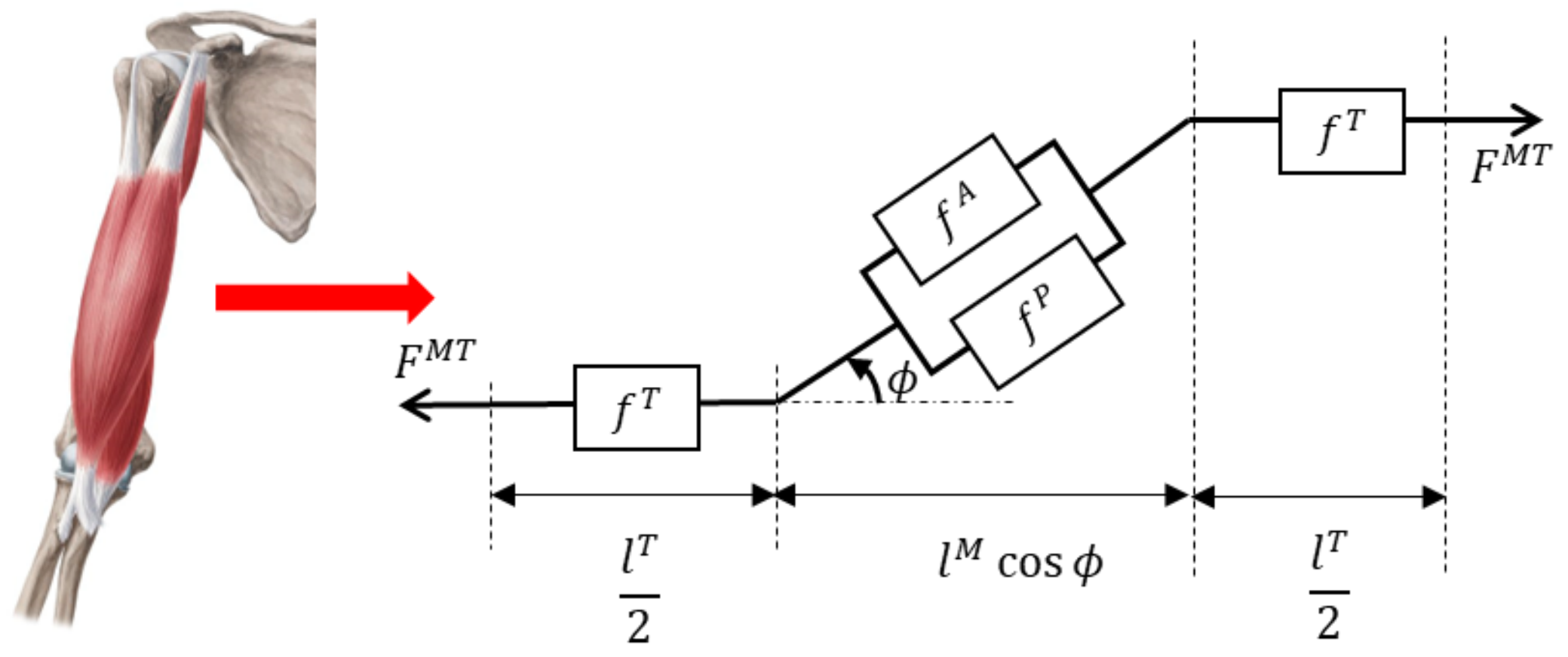

(a) (b)

**Fig. 2**: A muscle-tendon complex in the arm modelled by a homogenized hill-type model where muscle group's in (a) are a homogenized muscle-tendon (MT) complex described by the model shown in (b).

Forces in the muscle-tendon (MT) complex are generated by the dynamics of MT contractions, where for structural length scale behaviour of the MT complex, homogenized hill-type muscle models are utilized (described in Appendix B) . Each muscle group can be characterized by a parameter vector,

$$\boldsymbol{\kappa} = [l_0^M, v_{max}^M, f_0^M, l_s^T, \phi_0], \tag{3}$$

containing constants such as the maximum isometric force in the muscle ($f_0^M$), the optimal muscle length ($l_0^M$) corresponding to the maximum isometric force, the maximum contraction velocity ($v_{max}^{M}$), the slack length of the tendon ($l_s^T$), and the initial pennation angle ($\phi_0$) [56,57]. The total force produced by the MT complex, $F^{MT}$, can be expressed as:

$$F^{MT}(a, \tilde{l}^M, \tilde{v}^M, \phi; \boldsymbol{\kappa}) = F^M\big(a, \tilde{l}^M, \tilde{v}^M; \boldsymbol{\kappa}\big)\cos\phi. \tag{4}$$

where $a$ is the activation function in Eq. (2), $\tilde{l}^M$ is the normalized muscle length, $\tilde{v}^M$ is the normalized velocity of the muscle. In this study, the tendon is assumed to be rigid ($l^T = l_s^T$) which simplifies the MT contraction dynamics [58,59] accounting for the interaction of the activation, force length, and force velocity properties of the MT complex. More details can be found in Appendices A and B.

### 2.3 MSK Forward Dynamics of Motion

Body movement is the result of the force produced by actuators (MT complexes), converted to torques at the joints of the body, leading to rotation and translation of joints, which are considered as the generalized degrees of freedom of an MSK system ($\boldsymbol{q}$). The dynamic equilibrium can be expressed as

$$\boldsymbol{I}(\boldsymbol{q})\,\ddot{\boldsymbol{q}} - \boldsymbol{T}^{MT}(\boldsymbol{a}, \boldsymbol{q}, \dot{\boldsymbol{q}}; \boldsymbol{\kappa}) - \boldsymbol{E}(\boldsymbol{q}) = \boldsymbol{0}, \tag{5}$$

where $\boldsymbol{q}, \dot{\boldsymbol{q}}, \ddot{\boldsymbol{q}}$ are the vectors of generalized angular motions, angular velocities, and angular accelerations, respectively; $\boldsymbol{E}(\boldsymbol{q})$ is the torque from the external forces acting on the MSK system, e.g., ground reactions, gravitational loads etc.; $\boldsymbol{I}(\boldsymbol{q})$ is the inertial matrix; $\boldsymbol{T}^{MT}$ is the torque from all muscles in the model calculated by $\boldsymbol{T}^{MT}(\boldsymbol{a}, \boldsymbol{q}, \dot{\boldsymbol{q}}; \boldsymbol{\kappa}) = \boldsymbol{R}(\boldsymbol{q})\boldsymbol{F}^{MT}(\boldsymbol{a}, \boldsymbol{q}, \dot{\boldsymbol{q}}; \boldsymbol{\kappa})$, where $\boldsymbol{R}(\boldsymbol{q})$ are the moment arm's and $\boldsymbol{F}^{MT}(\boldsymbol{a}, \boldsymbol{q}, \dot{\boldsymbol{q}}; \boldsymbol{\kappa})$ are the forces from the MT complex. Given the muscle activation signals $\boldsymbol{a}$, initial conditions and parameters of involved muscle groups $\boldsymbol{\kappa}$, the generalized angular motions $\boldsymbol{q}$ and angular velocities $\dot{\boldsymbol{q}}$ of the joints can be obtained by solving Eq. (5). An example of these vectors is shown in Section 4 and Appendix D.

## 3. Multi-Resolution Recurrent Neural Networks for Physics-Informed Parameter Identification

This section describes the recurrent neural network algorithms, followed by the physics-informed parameter identification that enables the development of a forward dynamics surrogate and simultaneous parameter identification. The employment of multi-resolution analysis based on fast wavelet transform [60,61] for training data augmentation is then defined. The computational framework for multi-resolution recurrent neural network for physics-informed parameter identification is also discussed.

### 3.1 Recurrent Neural Networks and Gated Recurrent Units

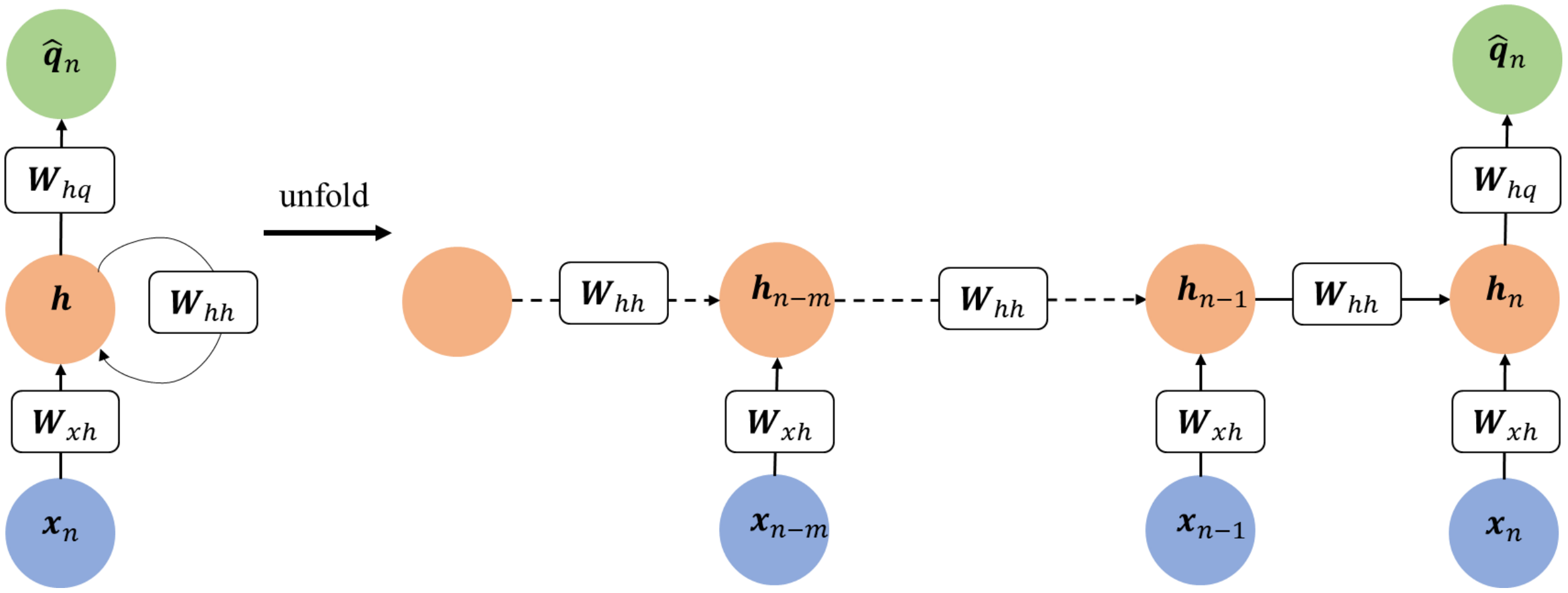

**Fig. 3**: Computational graph of a standard recurrent neural network using 'm' history steps for prediction.

The computational graph of a standard recurrent neural network (RNN) and its unfolded graph is shown in Fig. 3. The hidden state $\boldsymbol{h}$ allows for RNNs to learn important history-dependent features from the data in sequential time steps [29,53,54]. The unfolded graph shows the sharing of parameters across the architecture of the network, allowing for efficient training. The forward propagation of an RNN starts with an initial hidden state that embeds history-dependent features and propagates through all input steps. Considering an RNN with $m$ history steps as shown in Fig. 3, the propagation of the hidden state can be expressed as follows [29].

$$\boldsymbol{h}_i = a_{tanh}(\boldsymbol{W}_{hh}\boldsymbol{h}_{i-1} + \boldsymbol{W}_{xh}\boldsymbol{x}_i + \boldsymbol{b}_h), \quad i = n-m, \dots, n \tag{6}$$

The hidden state at the final (current) step $n$ is then used to inform the prediction.

$$\hat{\boldsymbol{q}}_n = \boldsymbol{W}_{hq}\boldsymbol{h}_n + \boldsymbol{b}_q \tag{7}$$

Here, $a_{tanh}$ is the hyperbolic tangent function; $\boldsymbol{W}_{xh}, \boldsymbol{W}_{hh},$ and $\boldsymbol{W}_{hq}$ are the trainable weight coefficients; $\boldsymbol{b}_h$ and $\boldsymbol{b}_q$ are the trainable bias coefficients. The trainable parameters are shared across all RNN steps. Let $\boldsymbol{x}_n = \left[t_n, e_n^1, \dots, e_n^{N_a}\right]$ be the current time and current sEMG data of the $N_a$ muscle components and $\widehat{\boldsymbol{q}}_n$ be the predicted joint motions at the current time $t_n$. Fig. 4(a) illustrates the computational graph of an RNN model trained to predict the motion at step $n$ by using *m* history steps of $\boldsymbol{x}$ and $\boldsymbol{q}$ as well as the $\boldsymbol{x}$ at step $n$. The forward propagation is defined as

$$\boldsymbol{h}_i = a_{tanh}\left(\boldsymbol{W}_{hh}\boldsymbol{h}_{i-1} + \boldsymbol{W}_{xh}\boldsymbol{x}_i + \boldsymbol{W}_{qh}\boldsymbol{q}_i + \boldsymbol{b}_h\right), i = n-m, \dots, n-1 \tag{8}$$

$$\boldsymbol{h}_n = a_{tanh}(\boldsymbol{W}_{hh}\boldsymbol{h}_{n-1} + \boldsymbol{W}_{xh}\boldsymbol{x}_n + \boldsymbol{b}_h), \tag{9}$$

$$\widehat{\boldsymbol{q}}_n = \boldsymbol{W}_{h\hat{q}}\boldsymbol{h}_n + \boldsymbol{b}_q \tag{10}$$

with trainable parameters including the weight coefficients $\boldsymbol{W}_{hh}, \boldsymbol{W}_{xh}, \boldsymbol{W}_{qh}$ and $\boldsymbol{W}_{h\hat{q}}$ and bias coefficients $\boldsymbol{b}_h$ and $\boldsymbol{b}_q$. During training, the ‘teacher-forcing’ method is used where the measured motion data is given to the model in the history steps. In test mode, the model is fed back to the previous predictions as input to inform future predictions. The inputs received in this scenario could be quite different from those passed through in the training process, leading the network to make extrapolative predictions and therefore, accumulate errors which will pollute the predictions. To improve the testing performance and enhance model accuracy and robustness, a user-controlled amount of random Gaussian noise is added to the recorded motion data to introduce stochasticity so that the network can learn variable input conditions, resembling those in the test mode, see [29] for details.

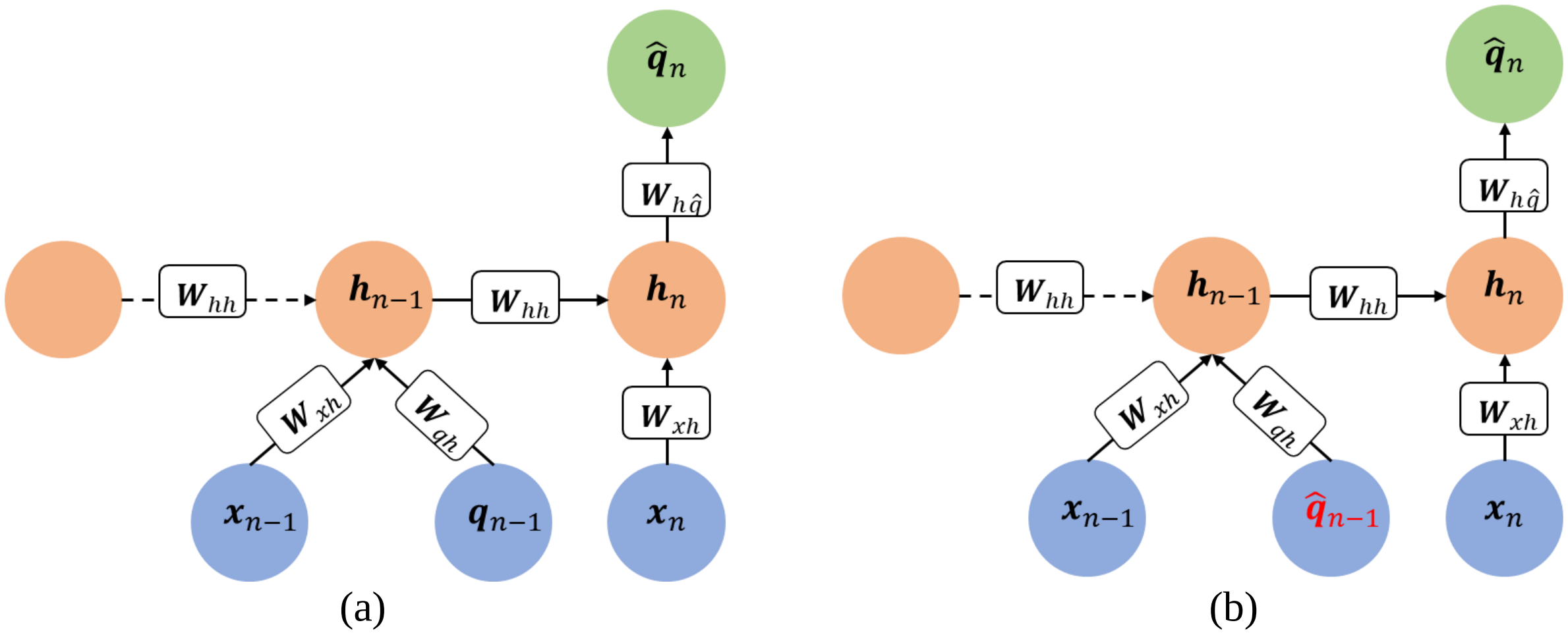

**Fig. 4**: An example computational graph of an RNN that uses one history step: (a) The train mode and (b) the test mode, where the motion predicted from the previous step is used as part of the input to predict motion at the current step.

Standard RNNs, however, have difficulties in learning long-term dependencies due to vanishing and exploding gradient issues arising from the recurrent connections. To mitigate these issues, gated recurrent units (GRUs) have been developed [48,50]. A standard GRU consists of a reset gate $\boldsymbol{r}_n$, that removes irrelevant history information, an update gate $\boldsymbol{u}_n$ that controls the amount of history information that is passed to the next step, and a candidate hidden state $\tilde{\boldsymbol{h}}_n$ that is used to calculate the current hidden state $\boldsymbol{h}_n$ [53,62]. Considering a GRU with $m$ history steps, the forward propagation can be expressed as follows [29]:

$$\begin{gathered}
\boldsymbol{r}_i = a_\sigma\left(\boldsymbol{W}_{hr}\boldsymbol{h}_{i-1} + \boldsymbol{W}_{xr}\boldsymbol{x}_i + \boldsymbol{W}_{qr}\boldsymbol{q}_i + \boldsymbol{b}_r\right) \\
\boldsymbol{u}_i = a_\sigma\left(\boldsymbol{W}_{hu}\boldsymbol{h}_{i-1} + \boldsymbol{W}_{xu}\boldsymbol{x}_i + \boldsymbol{W}_{qu}\boldsymbol{q}_i + \boldsymbol{b}_u\right) \\
\boldsymbol{z}_{(i,i-1)} = \boldsymbol{r}_i \odot \boldsymbol{W}_{h\tilde{h}}\boldsymbol{h}_{i-1} \\
\tilde{\boldsymbol{h}}_i = a_{tanh}\left(\boldsymbol{z}_{(i,i-1)} + \boldsymbol{W}_{x\tilde{h}}\boldsymbol{x}_i + \boldsymbol{W}_{q\tilde{h}}\boldsymbol{q}_i + \boldsymbol{b}_{\tilde{h}}\right) \\
\boldsymbol{c}_{(i,i-1)} = \boldsymbol{u}_i \odot \boldsymbol{h}_{i-1} \\
\tilde{\boldsymbol{c}}_{(i,i)} = \boldsymbol{u}_i \odot \tilde{\boldsymbol{h}}_i \\
\boldsymbol{h}_i = \boldsymbol{c}_{(i,i-1)} + \tilde{\boldsymbol{h}}_i - \tilde{\boldsymbol{c}}_{(i,i)} + \boldsymbol{b}_h \\
\\
\forall\, i = n-m, \dots, n-1,
\end{gathered} \tag{11}$$

$$
\begin{gathered}
\boldsymbol{r}_n = a_\sigma(\boldsymbol{W}_{hr}\boldsymbol{h}_{n-1} + \boldsymbol{W}_{xr}\boldsymbol{x}_n + \boldsymbol{b}_r) \\
\boldsymbol{u}_n = a_\sigma(\boldsymbol{W}_{hu}\boldsymbol{h}_{n-1} + \boldsymbol{W}_{xu}\boldsymbol{x}_n + \boldsymbol{b}_u) \\
\boldsymbol{z}_{(n,n-1)} = \boldsymbol{r}_n \odot \boldsymbol{W}_{h\tilde{h}}\boldsymbol{h}_{n-1} \\
\widetilde{\boldsymbol{h}}_n = a_{tanh}\left(\boldsymbol{z}_{(n,n-1)} + \boldsymbol{W}_{x\tilde{h}}\boldsymbol{x}_n + \boldsymbol{b}_{\tilde{h}}\right) \\
\boldsymbol{c}_{(n,n-1)} = \boldsymbol{u}_n \odot \boldsymbol{h}_{n-1} \\
\tilde{\boldsymbol{c}}_{(n,n)} = \boldsymbol{u}_n \odot \widetilde{\boldsymbol{h}}_n \\
\boldsymbol{h}_n = \boldsymbol{c}_{(n,n-1)} + \widetilde{\boldsymbol{h}}_n - \tilde{\boldsymbol{c}}_{(n,n)} + \boldsymbol{b}_h
\end{gathered} \tag{12}
$$

$$
\widehat{\boldsymbol{q}}_n = \boldsymbol{W}_{h\hat{q}}\boldsymbol{h}_n + \boldsymbol{b}_q \tag{13}
$$

where $\odot$ denotes the element-wise (Hadamard) product; $a_\sigma(\cdot)$ is the sigmoid activation function and $a_{tanh}(\cdot)$ is the hyperbolic tangent function; $\boldsymbol{W}_{hr}, \boldsymbol{W}_{xr}, \boldsymbol{W}_{qr}, \boldsymbol{W}_{hu}, \boldsymbol{W}_{xu}, \boldsymbol{W}_{qu}$, $\boldsymbol{W}_{h\tilde{h}}, \boldsymbol{W}_{x\tilde{h}}, \boldsymbol{W}_{q\tilde{h}}$ and $\boldsymbol{W}_{h\hat{q}}$ are the trainable weight coefficients; $\boldsymbol{b}_r, \boldsymbol{b}_u, \boldsymbol{b}_{\tilde{h}}, \boldsymbol{b}_h$ and $\boldsymbol{b}_q$ are the trainable bias coefficients. The current hidden state $\boldsymbol{h}_n$ is calculated by a linear interpolation between the previous hidden state $\boldsymbol{h}_{n-1}$ and the candidate hidden state $\widetilde{\boldsymbol{h}}_n$, based on the update gate $\boldsymbol{u}_n$. The model is trained via the backpropagation through time algorithm applied to RNNs [21]. Training occurs by plugging in the measured motion data in history steps (shown in Fig. 5), known as the teacher forcing procedure [21]. For predictions, the prediction from the previous step is used to predict the current step. The addition of gaussian noise to measured data, as described before, is adopted in GRU models as well.

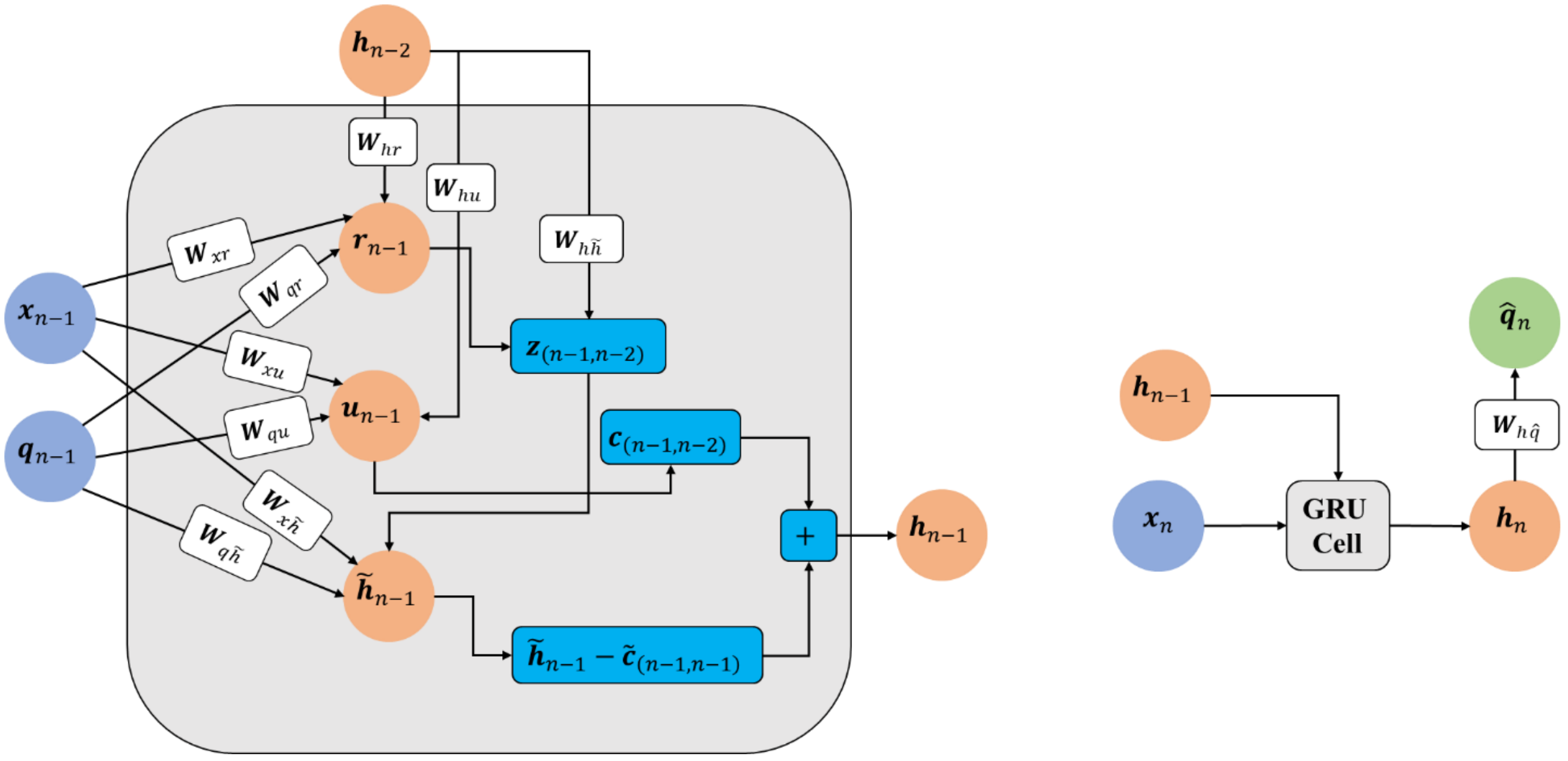

(a) (b)

**Fig. 5**: An example computational graph of a GRU in *train* mode that uses one history step: (a) Starting with an initial or previously obtained hidden state ($\boldsymbol{h}_{n-2}$), the main GRU cell takes the input $\boldsymbol{x}_{n-1}$ and motion $\boldsymbol{q}_{n-1}$ that are used to obtain the GRU hidden state $\boldsymbol{h}_{n-1}$ at step $n-1$ (Eq. (11)) and, (b) where the hidden state $\boldsymbol{h}_{n-1}$ is plugged back in to the GRU along with input $\boldsymbol{x}_n$ at step $n$ to predict the motion $\hat{\boldsymbol{q}}_n$ (Eq. (12)-(13)). The '+' cell produces an output (arrow pointing outwards) that is the summation of the inputs (arrows pointing into the cell).

### 3.2 Simultaneous Forward Dynamics Learning and Parameter Identification

With the governing equations for a general MSK forward dynamics (Section 2.1), the following parameterized ODE system is defined as

$$\boldsymbol{\mathcal{L}}[\boldsymbol{q}(t); \boldsymbol{\lambda}] = \boldsymbol{s}(t; \boldsymbol{\omega}),\ \forall\, t \in\ (0, \mathrm{T}],\ \boldsymbol{\mathcal{B}}[\boldsymbol{q}(0)] = \boldsymbol{g}\,, \tag{14}$$

where the differential operator $\boldsymbol{\mathcal{L}}[(\cdot); \boldsymbol{\lambda}]$ is parameterized by a set of parameters $\boldsymbol{\lambda}$. The right-hand side $\boldsymbol{s}(t; \boldsymbol{\omega})$ is parameterised by $\boldsymbol{\omega}$. $\boldsymbol{\mathcal{B}}[(\cdot)]$ is the operator for initial conditions, and $\boldsymbol{g}$ is the vector of prescribed initial conditions. To simplify notations, the ODE parameters are denoted by $\boldsymbol{\Gamma} = \{\boldsymbol{\lambda}, \boldsymbol{\omega}\}$. The solution to the ODE system $\boldsymbol{q}: [0, T] \rightarrow \mathbb{R}$ depends on the choice of parameters $\boldsymbol{\Gamma}$.

Here, an RNN is used to relate data inputs containing discrete sEMG signals and discrete time from all the $m$ previous history time-steps of a trial, $\cup_{i=n-m}^{n} \boldsymbol{x}_i \in \mathbb{R}^{n_{in}}, m \in \mathbb{Z}^+$, to discrete joint motion data outputs at the current time-step, $\boldsymbol{q}_n \in \mathbb{R}$ , approximating the MSK forward dynamics. Let the training input at the $i^{th}$ history step be defined as $\boldsymbol{x}_i = \left[t_i, e_i^1, \ldots, e_i^{N_a}\right]$, where $t_i$ denotes the time at the $i^{th}$ time step, and $\left\{e_i^j\right\}_{j=1}^{N_a}$ denotes the sEMG signals of $N_a$ muscle groups involved in the MSK joint motion at $t_i$. The motion at time step $n$, is then predicted using the training input from all the previous $m$ steps using the RNN.

$$\widehat{\boldsymbol{q}}_n(\boldsymbol{\theta}) = f_{RNN}(\boldsymbol{x}_n, \boldsymbol{x}_{n-1}, \boldsymbol{q}_{n-1}, \ldots, \boldsymbol{x}_{n-m}, \boldsymbol{q}_{n-m}; \boldsymbol{\theta}) \tag{15}$$

where $f_{RNN}$ denotes RNN evaluations (depending on model chosen) discussed in Eq. (11)-(13). The optimal RNN parameters $\widetilde{\boldsymbol{\theta}}$ and the ODE parameters $\widetilde{\boldsymbol{\Gamma}}$ are obtained by minimizing the composite loss function $J$ as follows,

$$\widetilde{\boldsymbol{\theta}}, \widetilde{\boldsymbol{\Gamma}} = \underset{\boldsymbol{\theta},\boldsymbol{\Gamma}}{\operatorname{argmin}}(J) = \underset{\boldsymbol{\theta},\boldsymbol{\Gamma}}{\operatorname{argmin}}(J_{data} + \beta J_{res}) \tag{16}$$

where $\beta$ is the parameter to regularize the loss contribution from the ODE residual term in the loss function and can be estimated analytically [1]. The data loss is defined by,

$$J_{data} = \frac{1}{N_{data}} \sum_{\alpha=1}^{N_{data}} \|\widehat{\boldsymbol{q}}_\alpha(\boldsymbol{\theta}) - \boldsymbol{q}_\alpha\|_{L_2}^2 \tag{17}$$

where $\widehat{\boldsymbol{q}}_\alpha(\boldsymbol{\theta})$ is the predicted motion, and $\boldsymbol{q}_\alpha$ is the recorded motion of MSK joints. In addition to training an MSK forward dynamics surrogate, the proposed framework aims to simultaneously identify important MSK parameters from the training data by minimizing residual of the governing equation of MSK system dynamics in Eq. (5).

$$\begin{gathered} J_{res} = \frac{1}{N_{data}} \sum_{\alpha=1}^{N_{data}} \|\boldsymbol{r}(\widehat{\boldsymbol{q}}_\alpha(\boldsymbol{\theta}); \boldsymbol{\Gamma})\|_{L_2}^2 \\ \boldsymbol{r}(\widehat{\boldsymbol{q}}_\alpha(\boldsymbol{\theta}); \boldsymbol{\Gamma}) = \boldsymbol{\mathcal{L}}[\widehat{\boldsymbol{q}}_\alpha(\boldsymbol{\theta}); \boldsymbol{\lambda}] - \boldsymbol{s}(t_\alpha; \boldsymbol{\omega}) \end{gathered} \tag{18}$$

where $\boldsymbol{r}(\widehat{\boldsymbol{q}}_\alpha(\boldsymbol{\theta}); \boldsymbol{\Gamma})$ is the residual associated with Eq. (14) for the $\alpha^{th}$ sample; $\boldsymbol{\Gamma} = \{\boldsymbol{\lambda}, \boldsymbol{\omega}\}$ represents the ODE parameters relevant to the MSK system. The gradients of the network outputs with respect to the network parameters ($\boldsymbol{\theta}$), MSK parameters ($\boldsymbol{\Gamma}$), and inputs are needed in the loss function minimization in Eq. (16), which can be obtained efficiently by automatic differentiation [63]. The formulation in Eq. (15) is general such that more advanced RNN frameworks can be used such as the GRU described in Eq. (11)-(13).

### 3.3 Multi-Resolution Training with Transfer Learning

To improve the training efficiency of RNN for MSK applications with mixed-frequency sEMG input signals and low-frequency output joint motion, a multi-resolution decomposition of the training input-output data is introduced in Section 3.3.1, followed by the transfer learning based multi-resolution training protocols to be discussed in Section 3.3.2.

#### 3.3.1 Wavelet based Multi-Resolution Analysis

Consider a sequence of nested subspaces $\ldots \subset V_{-1} \subset V_0 \subset V_1 \subset \cdots \subset L^2(R)$ where $\bigcup_{j\in\mathcal{Z}} V_j = L^2(R)$, and $\bigcap_{j\in\mathcal{Z}} V_j = \emptyset$. Each subspace $V_j$ of scale $[j]$ is spanned by a set of scaling functions $\phi_{j,k}(t)$, i.e.,

$$V_j = \left\{\phi_{j,k}(t) \middle| \phi_{j,k}(t) = 2^{\frac{j}{2}}\phi\left(2^j t - k\right), k \in \mathcal{Z}\right\}$$

Each subspace is related to the finer subspace through the law of dilation i.e., if $\phi(t) \in V_j$, then $\phi(2t) \in V_{j+1}, \forall j \in \mathcal{Z}$. Translations of the scaling function span the same subspace, i.e., if $\phi(t) \in V_j$, then $\phi(t-k) \in V_j, \forall j, k \in \mathcal{Z}$.

A mutually orthogonal complement of $V_j$ in $V_{j+1}$ is $W_j$, such that,

$$V_{j+1} = V_j \oplus W_j, \forall j \in \mathcal{Z} \tag{19}$$

where $\oplus$ is a direct sum. This subspace $W_j$ is spanned by a set of wavelet functions $\psi_{j,k}(t)$, i.e.,

$$W_j = \left\{\psi_{j,k}(t) \middle| \psi_{j,k}(t) = 2^{\frac{j}{2}}\psi\left(2^j t - k\right), k \in \mathcal{Z}\right\}$$

where $\psi(t)$ is the mother wavelet. It follows that,

$$\bigoplus_{j\in\mathcal{Z}} W_j = L^2(R) \tag{20}$$

and therefore,

$$V_j = V_i \oplus \left(\oplus_{k=0}^{j-i-1} W_{i+k}\right), j > i. \quad (21)$$

The two-scale dilation and translation relations for the scaling functions can be written as

$$\phi(t) = \sqrt{2} \sum_{k=-\infty}^{\infty} d_k \phi(2t - k). \quad (22)$$

Orthogonal wavelet functions can be obtained by imposing orthogonality conditions between scaling and wavelet functions in the frequency domain using Fourier transform,

$$\psi(t) = \sqrt{2} \sum_{k=-\infty}^{\infty} (-1)^{k-1} d_{-k-1} \phi(2t - k) \quad (23)$$

where $d_k$ is the coefficient.

Orthogonal scaling functions can be constructed by choosing a candidate function $\phi^*(t)$ such that $\phi^*(t)$ have reasonable decay and a finite support. In addition, $\int \phi^*(t)dt \neq 0$. It should also satisfy the two-scale relation,

$$\phi^*(t) = \sum_{k} p_k \phi^*(2t - k), k \in \mathcal{Z}. \quad (24)$$

With these, an orthogonal scaling function $\phi(t)$ can be expressed in terms of $\phi^*(t)$ as

$$\phi(t) = \sum_{k=-\infty}^{\infty} a_k \phi^*(t - k). \quad (25)$$

It is then possible to define the scaling function at the coarse scale in terms of the scaling function at the fine scale and the wavelet functions at the coarser scale,

$$\phi(2t - l) = \sum_{k=-\infty}^{\infty} d_{l-2k} \phi(t - k) + \sum_{k=-\infty}^{\infty} h_{l-2k} \psi(t - k), l \in \mathcal{Z}. \quad (26)$$

Any function can be approximated at scale $[j]$ by using $\phi_{j,k}$ as a basis as well as using its coarse scale $[j-1]$ representation and details at the coarse scale, i.e.,

$$P_j f = \sum_{k=-\infty}^{\infty} S_k^{[j]} \phi_{j,k} = P_{j-1} f + H_{j-1} f \tag{27}$$

$$= \sum_{k=-\infty}^{\infty} S_k^{[j-1]} \phi_{j-1,k} + \sum_{k=-\infty}^{\infty} T_k^{[j-1]} \psi_{j-1,k}$$

where $P_j$ and $H_j$ are the operators projecting $f$ onto the subspaces $V_j$ and details of $f$ at scale $[j]$ in the orthogonal subspace $W_j$, respectively. $S_k^{[j]}$ and $T_k^{[j]}$ are the corresponding basis coefficients at the coarse scale $[j]$. While the example shown here is for a one-dimensional case, this multi-resolution representation can be extended to multi-dimensions.

### 3.3.2 Multi-Resolution Data Representation and Training Protocols

In this approach, a given signal $f(t)$ is represented using the multi-resolution scaling functions and wavelets. A scale $[j]$ representation of signal $f(t)$ can be obtained from the scale $[r]$ $(j > r)$ representation with the addition of wavelet components (high frequency components) of the scales higher than $[r]$, using the discrete wavelet transform modified from Eq. (27),

$$P_j f(t) = P_r f(t) + \sum_{b=r}^{j-1} H_b f(t) = \sum_{k=-\infty}^{\infty} S_k^{[r]} \phi_{r,k}(t) + \sum_{b=r}^{j-1} \sum_{k=-\infty}^{\infty} T_k^{[b]} \psi_{b,k}(t) \tag{28}$$

where $P_r$ is the projection operator at scale $[r]$ and $H_b$ are the wavelet projectors of the signal that are added from scale $[r]$ to scale $[j-1]$ to reconstruct the signal at scale $[j]$; $S_k^{[r]}$ and $T_k^{[b]}$ are the scaling and wavelet function's coefficients, obtained by the orthogonality condition as given in Section 3.3.1.

Using the Wavelet transform to represent a time series under multiple resolutions offers advantages for feature extraction from signals. Compared to the Fourier transform which offers only localization in the frequency domain, the Wavelet transform provides both frequency and time domain localization, making it more suitable for time history (or sequence) learning algorithms such as the standard RNN and its enhanced variant GRU. More specifically, one can enhance training efficiency by using a sequential training strategy for the time-history input (sEMG) and output (joint motion) data. Applying the Fast Wavelet Transform [57,58] to obtain the input and output data from low to high resolutions results in better generalization performance of the RNN trained to map from sEMG signals to joint motion time history as described below. The second order Daubechies wavelets are used in this work.

Here we consider a general MSK system described in Section 2. The original unfiltered data is denoted as scale [0], which will be decomposed into a sequence of lower scales $[-j], j \in \mathbb{Z}^+$ for multi-resolution training.

Let $\boldsymbol{D}^{[0]}$ be the input training data at the full-scale $(j = 0)$ of the raw signals i.e.,

$$\boldsymbol{D}^{[0]} = \left[\boldsymbol{x}_1^{[0]}, \boldsymbol{x}_2^{[0]}, \dots, \boldsymbol{x}_{N_{data}}^{[0]}\right], \qquad (29)$$
$$\boldsymbol{x}_i^{[0]} = \left[t_i, e_i^{1[0]}, \dots, e_i^{N_a[0]}\right].$$

and the motion of joints of the MSK system at the $i^{th}$ time-step at the full-scale $(j = 0)$ is $\boldsymbol{q}_i^{[0]}$ such that the array of the unfiltered motion data for the duration of the motion is $\boldsymbol{q}^{[0]} = \left[\boldsymbol{q}_1^{[0]}, \boldsymbol{q}_2^{[0]}, \dots \boldsymbol{q}_{N_{data}}^{[0]}\right]$.

From MR theory, subtracting details from the fine scale representations at the full-scale of the signal, i.e., [0], results in a course scale representation of the signal at scale $[-k], k = 1, \dots j$. The projected training data at coarse scale [-$j$] is defined as

$$\boldsymbol{D}^{[-j]} = \left[\boldsymbol{x}_1^{[-j]}, \boldsymbol{x}_2^{[-j]}, \dots, \boldsymbol{x}_{N_{data}}^{[-j]}\right], \qquad (30)$$

where $N_{data}$ is the total number of data points and

$$\boldsymbol{x}_i^{[-j]} = \left[t_i, e_i^{1[-j]}, \dots, e_i^{N_a[-j]}\right], i\text{=}1\dots, N_{data} \qquad (31)$$

is the input data of scale [-$j$] at time step $i$. The motion of the MSK joints at the $i^{th}$ time-step at the scale $[-j]$ is $\boldsymbol{q}_i^{[-j]}$. The data sets for a representative muscle group '$MT$', $e_i^{MT[-j]}$ and motion $\boldsymbol{q}_i^{[-j]}$, are obtained from the original raw data $e_i^{MT[0]}$ and $\boldsymbol{q}_i^{[0]}$ by wavelet projection using Eq. (27), that is,

$$\begin{aligned} e^{MT[-j]}(t) &\equiv P_j e^{MT[0]}(t) = P_{j-1} e^{MT[0]}(t) + H_{j-1} e^{MT[0]}(t) \\ &= e^{MT[0]}(t) - \sum_{b=0}^{j-1} H_b e^{MT[0]}(t) \end{aligned} \qquad (32)$$

$$\boldsymbol{q}^{[-j]}(t) \equiv \boldsymbol{P}_j \boldsymbol{q}^{[0]}(t) = \boldsymbol{P}_{j-1} \boldsymbol{q}^{[0]}(t) + \boldsymbol{H}_{j-1} \boldsymbol{q}^{[0]}(t) = \boldsymbol{q}^{[0]}(t) - \sum_{b=0}^{j-1} \boldsymbol{H}_b \boldsymbol{q}^{[0]}(t)$$

where $\boldsymbol{P}_j$ and $\boldsymbol{H}_j$ are the projection operators in multi-dimensions. Hence, datasets that contain lower resolution representations of the original signal at scales [0] can be expressed as:

$$\boldsymbol{D}^{[-j]} \subset \boldsymbol{D}^{[-j+1]} \subset \cdots \boldsymbol{D}^{[-1]} \subset \boldsymbol{D}^{[0]} \tag{33}$$
$$\boldsymbol{q}^{[-j]} \subset \boldsymbol{q}^{[-j+1]} \subset \cdots \boldsymbol{q}^{[-1]} \subset \boldsymbol{q}^{[0]}$$

where $\boldsymbol{q}^{[-j]} = \left[\boldsymbol{q}_1^{[-j]}, \boldsymbol{q}_2^{[-j]}, \dots, \boldsymbol{q}_{N_{data}}^{[-j]}\right]$.

Instead of learning the signal mapping from input original raw sEMG data $\boldsymbol{D}^{[0]}$ to motion data $\boldsymbol{q}^{[0]}$, we initiate learning the mapping by starting from a *coarse* scale representation of the input-output data at scale $[-j]$ and map $\boldsymbol{D}^{[-j]}$ to $\boldsymbol{q}^{[-j]}$. For multi-resolution RNN, the initial learning starts from the coarsest scale $[-j]$ as follows:

$$\boldsymbol{h}_i^{[j]} = a_{tanh}\left(\boldsymbol{W}_{hh}^{[-j]} \boldsymbol{h}_{i-1}^{[-j]} + \boldsymbol{W}_{xh}^{[-j]} \boldsymbol{x}_i^{[-j]} + \boldsymbol{W}_{qh}^{[-j]} \boldsymbol{q}_i^{[-j]} + \boldsymbol{b}_h^{[-j]}\right), \tag{34}$$
$$\forall i = n-m, \dots, n-1$$
$$\boldsymbol{h}_n^{[-j]} = a_{tanh}\left(\boldsymbol{W}_{hh}^{[-j]} \boldsymbol{h}_{n-1}^{[-j]} + \boldsymbol{W}_{xh}^{[-j]} \boldsymbol{x}_n^{[-j]} + \boldsymbol{b}_h^{[-j]}\right), \tag{35}$$

$$\widehat{\boldsymbol{q}}_n^{[-j]} = \boldsymbol{W}_{h\hat{q}}^{[-j]} \boldsymbol{h}_{n-1}^{[-j]} + \boldsymbol{b}_q^{[-j]}. \tag{36}$$

At the next finer scale $[-j+1]$, the weights at scale $[-j]$ (using an early stopping [64]) are used as the initial values for $\boldsymbol{W}_{hh}^{[-j+1]}$, $\boldsymbol{W}_{xh}^{[-j+1]}$, $\boldsymbol{W}_{qh}^{[-j+1]}$, $\boldsymbol{W}_{h\hat{q}}^{[-j+1]}, \boldsymbol{b}_h^{[-j]}, \boldsymbol{b}_q^{[-j]}$, similar to the concept of transfer learning[65].

Similarly, for multi-resolution GRU, the initial learning starts from the coarsest scale $[-j]$ as described in Appendix C. The same procedures to transfer the NN parameters in Eq.(34)-(36) are repeated with $[-j] \rightarrow [-j+1]$ until it reaches scale [0]. To enhance model accuracy and robustness, variations based on Gaussian noise are added to the motion data in each sequential step, as suggested by [29]. The sequential MR training process is described in Algorithm 1.

Algorithm 1: Sequential Multi-Resolution PI-RNN training process.

Step1: Initialize parameters.

$$\boldsymbol{\theta} = \boldsymbol{\theta}^0, \boldsymbol{\Gamma} = \boldsymbol{\Gamma}^0$$

Step 2: Sequential learning through parameter transfer from coarse-scale to fine-scale.

For $l = 0 \rightarrow j$

1. Train the RNN on the dataset $\boldsymbol{D}^{[-j+l]}$ by calculating the predictions as

$$\widehat{\boldsymbol{q}}_n^{[-j+l]}(\boldsymbol{\theta}, \boldsymbol{\Gamma}) = f_{RNN}\left(\boldsymbol{x}_n^{[-j+l]}, \boldsymbol{x}_{n-1}^{[-j+l]}, \boldsymbol{q}_{n-1}^{[-j+l]}, \dots, \boldsymbol{x}_{n-m}^{[-j+l]}, \boldsymbol{q}_{n-m}^{[-j+l]}; \boldsymbol{\theta}, \boldsymbol{\Gamma}\right)$$

$m \rightarrow$ # of history steps

$n \rightarrow$ current time step

$$\widetilde{\boldsymbol{\theta}}, \widetilde{\boldsymbol{\Gamma}} = \underset{\boldsymbol{\theta}, \boldsymbol{\Gamma}}{\operatorname{argmin}}(J) = \underset{\boldsymbol{\theta}, \boldsymbol{\Gamma}}{\operatorname{argmin}}\left(J_{data}(\boldsymbol{\theta}) + \beta\, J_{res}(\boldsymbol{\theta}, \boldsymbol{\Gamma})\right)$$

2. $\boldsymbol{\theta} = \widetilde{\boldsymbol{\theta}}, \boldsymbol{\Gamma} = \widetilde{\boldsymbol{\Gamma}}$

## 4. Verification Example

For verification of the proposed MR PI-RNN framework, an elbow flexion-extension model [1] and synthetic sEMG signals with gaussian noise and associated motion responses were considered. The flowchart of the proposed computational framework for simultaneous forward dynamics prediction and parameter identification of MSK parameters is shown in Fig. 6.

The model contained two rigid links corresponding to the upper arm and forearm with lengths $l_{ua}$ and $l_{fa}$, respectively. They were connected at a hinge resembling the elbow joint “A”, while the upper arm link was fixed at the top joint “B”, and the biceps (*Bi*) and triceps (*Tri*) muscle-tendon complexes (modeled by Hill-type models with parameters $\boldsymbol{\kappa}_{Bi}$ and $\boldsymbol{\kappa}_{Tri}$) were represented by the lines connecting the links, as shown in Fig. 6. The degree of freedom of the model was the elbow flexion angle $q$. The mass in the forehand was assumed to be concentrated at the wrist location, hence, a mass $m_{fa}$ was attached to one end of the forearm link with a moment arm $l_{fa}$ from the elbow joint. Tendons were assumed as rigid [58] for ease of computation.

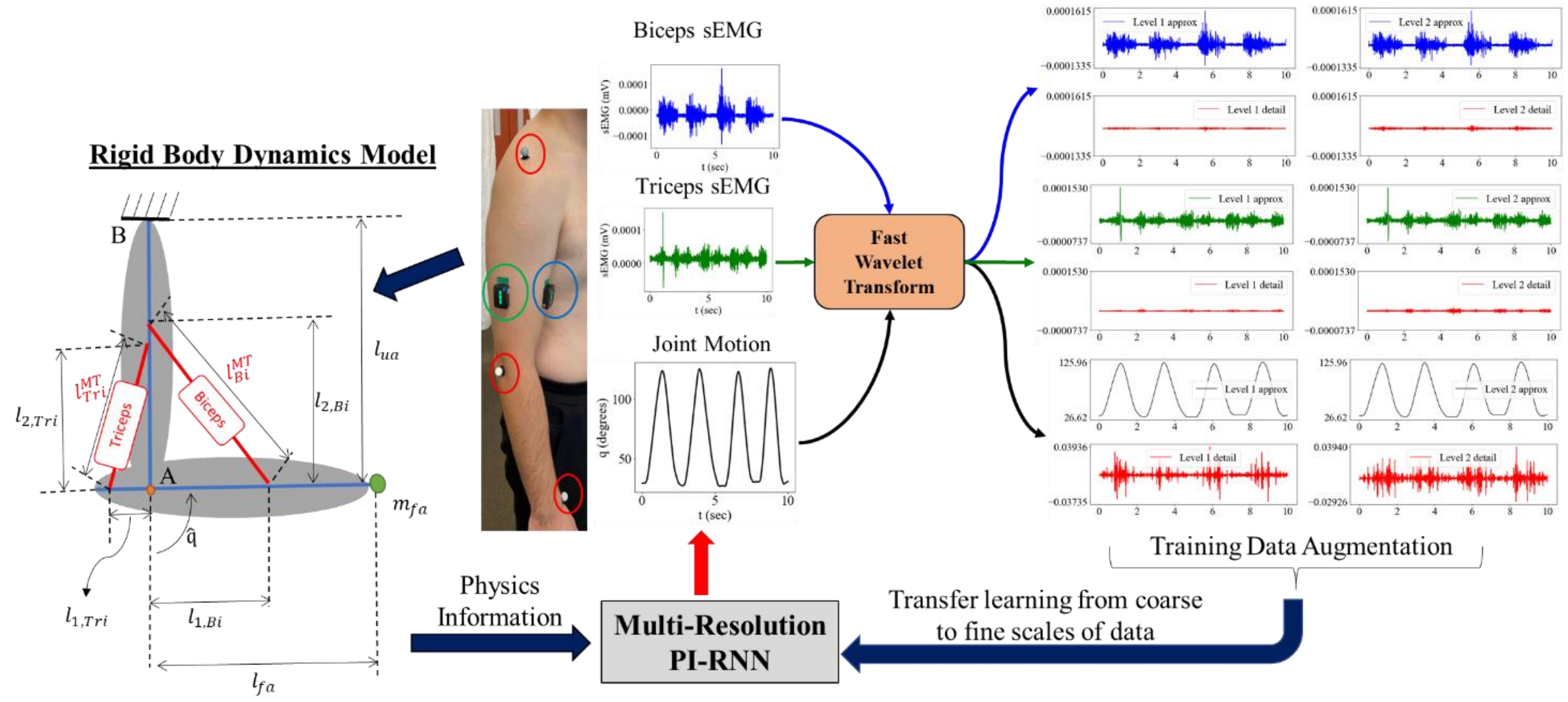

**Fig. 6**: An overview of the application of this framework to the recorded motion data. The location of motion capture markers is circled in red and the sEMG sensors on Biceps and Triceps muscle groups in blue and green, respectively. The simplified rigid body model was used in the forward dynamics equations within the framework with appropriately scaled anthropometric properties (for geometry) and physiological parameters (for muscle-tendon material models). The raw sEMG signals were mapped to the target angular motion of the elbow and used to simultaneously characterize the MSK system using the proposed Multi-Resolution PI-RNN framework.

The equation of motion for this rigid body system is given in Appendix D. Given the synthetic sEMG signals $(e^{Bi}(t), e^{Tri}(t))$, the initial conditions $q(0) = \frac{\pi}{6}$ radians and $\dot{q}(0) = 0$ radians/sec and the parameters in Table 1, the motion of the elbow joint, $q$, can be obtained by solving the MSK forward dynamics problem using a synthetic solver.

Table 1: Parameters involved in the forward dynamics setup of elbow flexion-extension motion.

| Parameter | Type | Value | Parameter | Type | Value |
|---|---|---|---|---|---|
| $l_{0,\mathrm{Bi}}^{M}$ | Biceps Muscle Model | 0.6 m | $m_{fa}$ | Equation of motion | 1.0 kg |
| $v_{max,Bi}^{M}$ | Biceps Muscle Model | 6 m/sec | $l_{ua}$ | Geometric | 1.0 m |
| $f_{0,\mathrm{Bi}}^{M}$ | Biceps Muscle Model | 300 N | $l_{fa}$ | Geometric | 1.0 m |
| $l_{s,Bi}^{T}$ | Biceps Muscle Model | 0.55 m | $l_{1,Bi}$ | Geometric | 0.3 m |
| $\phi_{Bi}$ | Biceps Muscle Model | 0.0 radians | $l_{2,Bi}$ | Geometric | 0.8 m |
| $l_{0,\mathrm{Tri}}^{M}$ | Triceps Muscle Model | 0.4 m | $l_{1,Tri}$ | Geometric | 0.2 m |
| $v_{max,Tri}^{M}$ | Triceps Muscle Model | 4 m/sec | $l_{2,Tri}$ | Geometric | 0.7 m |
| $f_{0,Tri}^{M}$ | Triceps Muscle Model | 300 N | $d$ | Activation Dynamics | 0.08 sec |
| $l_{s,Tri}^{T}$ | Triceps Muscle Model | 0.33 m | A | Activation Dynamics | 0.2 |
| $\phi_{Tri}$ | Triceps Muscle Model | 0.0 radians | | | |

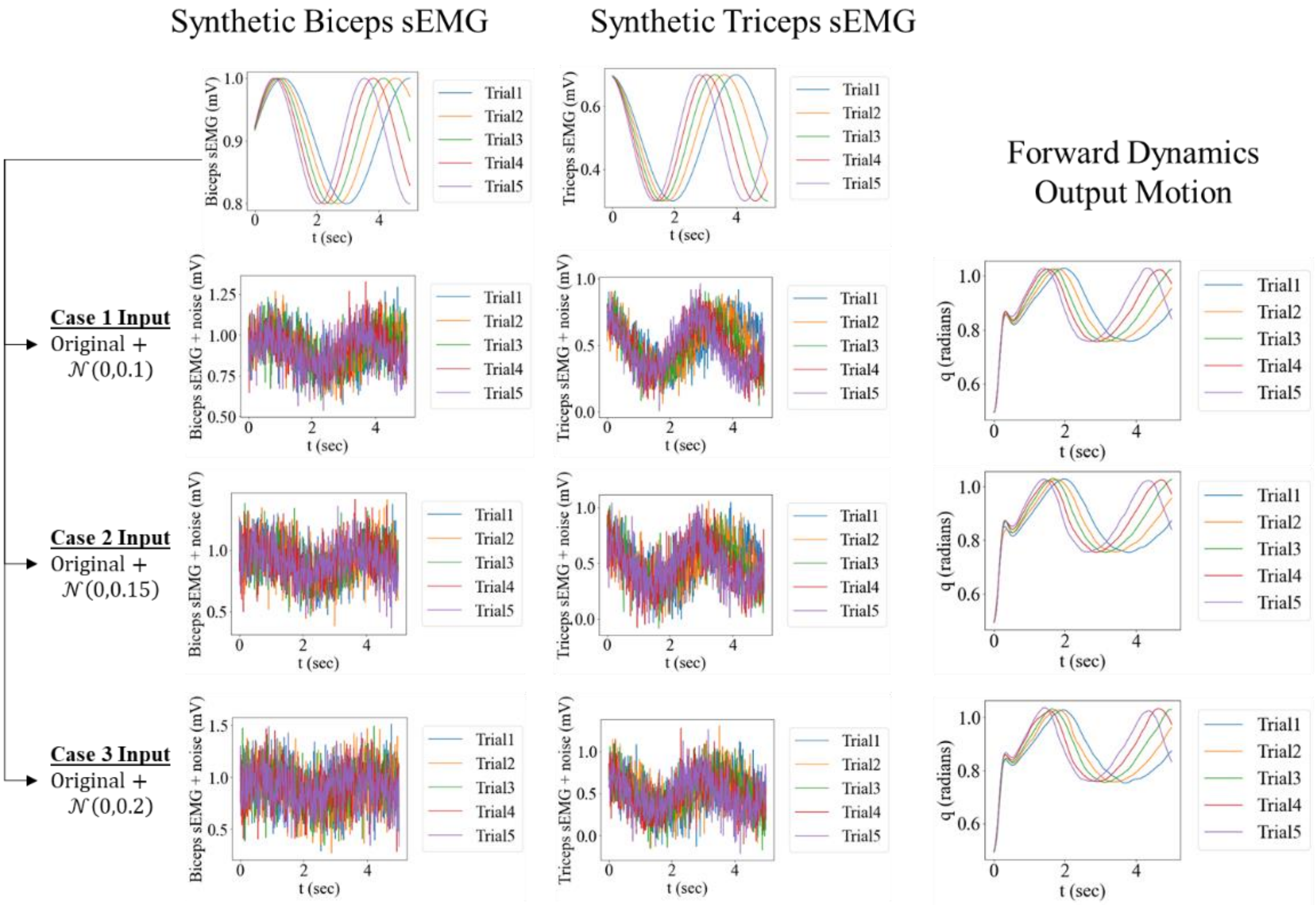

**Fig. 7**: The original 'noiseless' input data set with the synthetic biceps and triceps sEMG signals having variations in frequency for five trials are shown at the top. Increasing levels of noise are added to develop 3 cases of synthetic mixed frequency input sEMG, from which corresponding output motions are solved, using the forward dynamics equations. To verify the MR framework, these three cases with their respective mixed frequency input data are then mapped to their corresponding motion data.

To verify and check the robustness of the MR framework to different levels of noise in the input, the following test was performed. Originally, five synthetic samples i.e., Trial's 1 to 5, of *noiseless* synthetic muscle sEMG signals are assumed, as shown in Fig. 7. In practical applications, signals obtained from measurement devices such as sEMG sensors contain noise in their content. Therefore, three cases were developed by adding Gaussian noise $(\mathcal{N}(\mu, \sigma))$ with zero mean $(\mu = 0)$ and increasing levels of standard deviations $(\sigma)$ to the input synthetic sEMG signals as mentioned in Table 2. As the maximum value of the *noiseless* sEMG signals is 1, the chosen $\sigma$'s were kept within 10% - 20% of the signal maximum for a reasonable level of noise. Then the corresponding output motions are generated by passing the noisy sEMG as input to the FD equations in Section 2. The following training procedures were performed for each of the three cases.

**1-scale Training**

The mixed frequency input sEMG signals and corresponding output motion data $\boldsymbol{q}$ at scale [0], denoted by $\boldsymbol{D}^{[0]}$ and $\boldsymbol{q}^{[0]}$, respectively, are mapped to get a baseline performance. This is termed as *1-scale* training as only the full-scale (i.e., [0]) of the mixed frequency data is used for training.

**2-scale Training**

a. Initiate learning from a *coarse* scale representation of the mixed frequency input data at scale $[-1]$ and map $\boldsymbol{D}^{[-1]}$ to the corresponding motion data at scale [-1], $\boldsymbol{q}^{[-1]}$**,** of that case.
b. Transfer parameters to the next scale training and finish the learning by mapping $\boldsymbol{D}^{[0]}$ to $\boldsymbol{q}^{[0]}$.

**3-scale Training**

a. Start learning from a *coarse* scale representation of the mixed frequency input data at scale $[-2]$ and map $\boldsymbol{D}^{[-2]}$ to the corresponding motion data at scale [-2] , $\boldsymbol{q}^{[-2]}$**,** of that case.
b. Transfer parameters to the next scale training and continue learning by mapping $\boldsymbol{D}^{[-1]}$ to $\boldsymbol{q}^{[-1]}$.
c. Transfer parameters to the next scale training and finish the learning by mapping $\boldsymbol{D}^{[0]}$ to $\boldsymbol{q}^{[0]}$.

Table 2: Input data and gaussian noise level for each case.

| **Case ID** | **Input Synthetic sEMG data** + $\mathcal{N}(\mu,\sigma)$ |
|---|---|
| 1 | Original +$\mathcal{N}(0,0.1)$ |
| 2 | Original +$\mathcal{N}(0,0.15)$ |
| 3 | Original +$\mathcal{N}(0,0.2)$ |

For each case and for each of the training scales in that case, the training data samples contained the data of trial's 1, 2, 4, 5 while trial 3 was used for testing, each trial with $n$ =500 data points. The MSK parameters $\boldsymbol{\Gamma} = \{\Gamma_l\}_{l=1}^{4} = \left\{f_{0,Bi}^{M}, l_{0,Bi}^{M}, f_{0,Tri}^{M}, l_{0,Tri}^{M}\right\}$ were chosen to be identified from the training data using the proposed framework. Due to differences in units and physiological nature of the parameters, the conditioning of the parameter identification system could be affected. To mitigate this issue, normalization [1,44] was applied to each of the parameters,

$$\bar{\Gamma}_l = \frac{\Gamma_l}{\Gamma_l^{(0)}} \tag{37}$$

where $\Gamma_l^{(0)}$ was the initial value of the parameter. Therefore, the parameters to be identified became $\bar{\boldsymbol{\Gamma}} = \{\bar{\Gamma}_l\}_{l=1}^{4}$.

The proposed framework, as described in Section 3, was applied to each case to simultaneously learn the MSK forward dynamics surrogate and identify the MSK parameters $\bar{\boldsymbol{\Gamma}}$ by optimizing Eq. (16), where the residual of the governing equation for the current time step $k$, was expressed as

$$\begin{aligned} & r\left(\hat{q}_k^{[-j]}(\boldsymbol{\theta_q}), \dot{\hat{q}}_k^{[-j]}(\boldsymbol{\theta_q}), \ddot{\hat{q}}_k^{[-j]}(\boldsymbol{\theta_q}); \boldsymbol{\Gamma}(\bar{\boldsymbol{\Gamma}}; \boldsymbol{\Gamma}^{(0)})\right) \\ & \quad = I\ddot{\hat{q}}_k^{[-j]}(\boldsymbol{\theta_q}) - E\left(\hat{q}_k^{[-j]}(\boldsymbol{\theta_q})\right) \\ & \qquad\qquad - T^{MT}\left(a_{Bi}(t_k), a_{Tri}(t_k), \hat{q}_k^{[-j]}(\boldsymbol{\theta_q}), \dot{\hat{q}}_k^{[-j]}(\boldsymbol{\theta_q}); \boldsymbol{\Gamma}(\bar{\boldsymbol{\Gamma}}; \boldsymbol{\Gamma}^{(0)})\right) \end{aligned} \tag{38}$$

and is included in the residual term $J_{res}$ in the loss function in Eq. (16). While the training happens sequentially from coarse to fine-scales of the motion, the final identification of parameters happens at the scale [0], i.e., the full-scale in each of the 1-, 2- and 3-scale MR training types.

A GRU with 2 history steps, 1 hidden layer and 50 neurons in each layer was used. The training was performed using the Adam algorithm [66] with an initial learning rate of $1 \times 10^{-3}$ and the penalty parameter for the MSK residual term in the loss function, $\beta \propto \frac{\Delta t^2}{I} = 10^{-3}$. $\Delta t$ is the time-step between data points and $I$ is the moment of inertia in Eq. (38). Multiple parameter initialization seeds were used for an averaged response of the MR training.

To compare the post-training performance of 1-, 2- and 3-scale MR training's, the average testing mean squared error (MSE) and testing $\mathrm{R}^2$ scores were compared, where these measures for a single trial are defined as:

$$\mathrm{MSE} = \frac{1}{n}\|\boldsymbol{q} - \hat{\boldsymbol{q}}\|_{L_2}^2 \tag{39}$$

$$\mathrm{R}^2 = 1 - \frac{\sum_{i=1}^{n}(q_i - \hat{q}_i)^2}{\sum_{i=1}^{n}(q_i - \bar{q})^2} \tag{40}$$

where $\boldsymbol{q}$ is the motion data of the trial, $\hat{\boldsymbol{q}}$ is the trial's predicted motion from the MR PI-RNN framework, and $\bar{q}$ is the mean of trial's motion data with $n$ being the number of data points in the trial. At each epoch in the MR training, the *training loss* is calculated by using the scale of the *training data* used in that training scale, i.e., scale $[-j]$ of the data is used in $j$-scale training.

The gradual improvement in these metrics is evident from Fig. 8 where, as further scales of information are added and the training data is augmented, the generalization performance shows improvement from 1-scale to 3-scale. Overall, it is noted that the test metrics such as the MSE reduces, and the $\mathrm{R}^2$ score gets closer to one, indicating an increase in the generalization accuracy as more training scales are introduced. This can be explained through the theory of bias-variance tradeoff; training on various scales of the data introduces more variance to the training, helping the ML framework to reduce the bias it develops by just training on the full-scale of the data. Together, this reduction in bias and growth in variance leads to a better generalization performance. Computationally, this method improves accuracy in the same amount of training epochs showing the efficiency of this method. As generalization predictions post-training are made using the full-scale of the data, there is no increase in time needed to perform the forward pass for any scale.

Meanwhile, the MSK parameters, $f_0^M$ (maximum isometric force) and $l_0^M$ (optimal muscle length corresponding to the maximum isometric force), of both the biceps and the triceps were accurately identified from the motion data, as shown in Table 3. Compared with the parameter identification from our previous work [1] where in addition to $f_0^M$, the maximum contraction velocity $v_{max}^M$ was independently identified, due to non-convergence of $l_0^M$ by the time-domain and feature-encoded trainings, the proposed method can accurately identify $l_0^M$. $v_{max}^M$ can then by obtain by the experimentally observed relationship of $v_{max}^M / l_0^M = 10s^{-1}$ [58,67].

For the identification of optimal muscle length parameters $(l_0^M)$, the initial points need to be chosen with respect to constraints applied by the geometry of the MSK system. The errors reported in Table 3 are calculated by taking the average of the percentage error of the identified MSK parameters from the 3-scale training with the multiple parameter $(\boldsymbol{\theta}, \boldsymbol{\Gamma})$ initializations. It was observed that in MSK parameter identification, similar accuracy in characterization was obtained from all training scale approaches used within each case, with errors less than 1%. This indicates that the MR PI-RNN improves the generalization performance of the motion prediction, without loss in parameter identification accuracy.

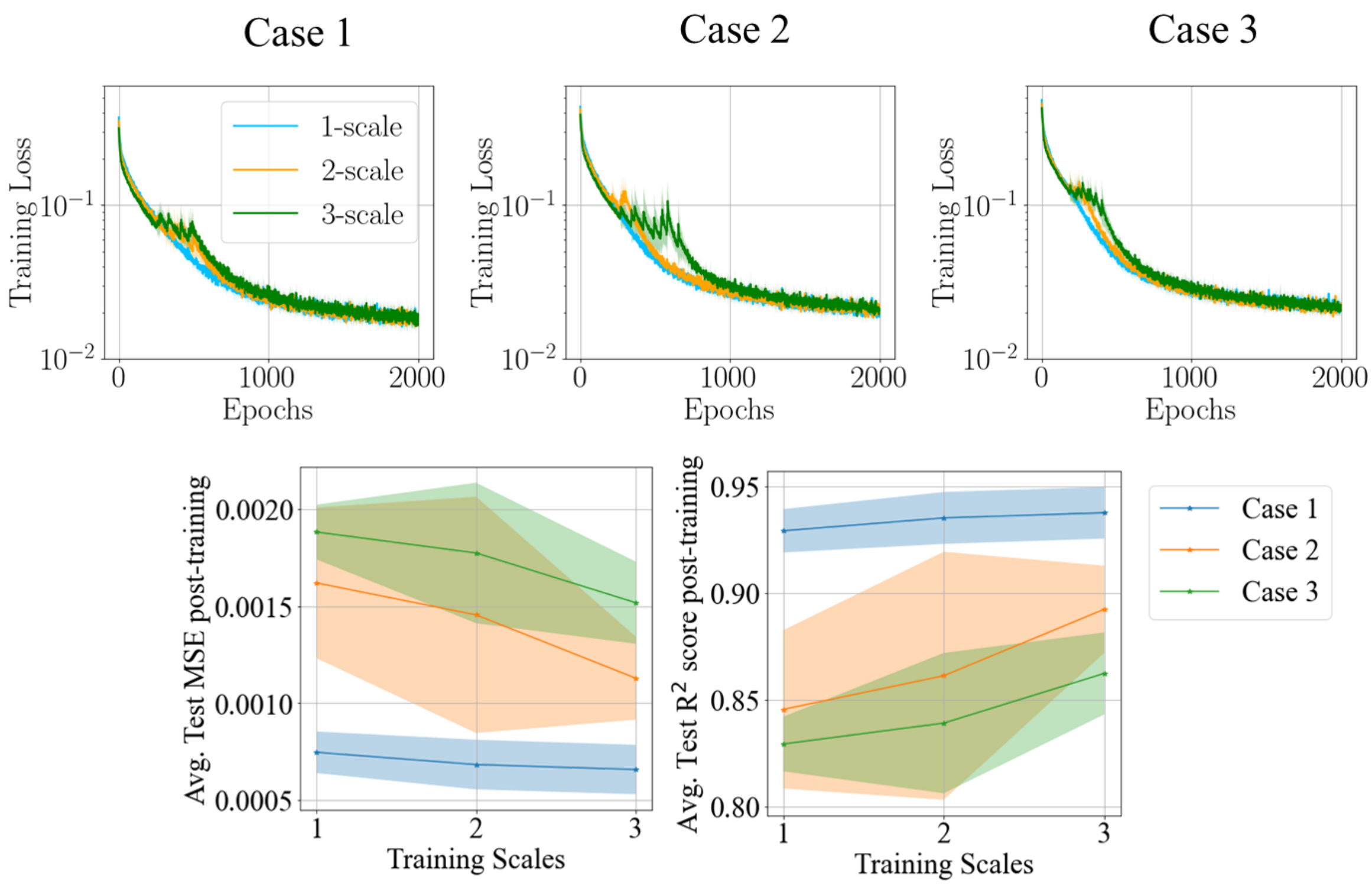

**Fig. 8**: The training loss and testing metrics are shown. The shaded area in the loss and average test MSE and $R^2$ score figures is one standard deviation from the mean (solid line). As more scales of data are introduced in the MR training, the average Test MSE and $R^2$ score calculated post-training improve in each case.

Table 3: The average percentage error (shown as mean $\pm$ standard deviation) between predicted and true values of the parameters for 3-scale training for each case from multiple initialization points.

| **Parameter** | **Case 1** | **Case 2** | **Case 3** |
|---|---|---|---|
| $f_{0,Bi}^{M}$ | $0.50 \pm 0.02$ | $0.50 \pm 0.01$ | $0.37 \pm 0.04$ |
| $f_{0,Tri}^{M}$ | $0.06 \pm 0.02$ | $-0.04 \pm 0.02$ | $-0.02 \pm 0.03$ |
| $l_{0,Bi}^{M}$ | $0.10 \pm 0.03$ | $0.10 \pm 0.02$ | $0.05 \pm 0.05$ |
| $l_{0,Tri}^{M}$ | $-0.06 \pm 0.07$ | $-0.05 \pm 0.02$ | $-0.03 \pm 0.10$ |

# 5. Validation: Elbow Flexion-Extension Motion

## 5.1 Application of MR PI-RNN method to Subject-Specific Data

The recorded motion data and sEMG signals were collected and processed as per the data acquisition protocols mentioned in [1]. Three elbow flexion-extension motion trials were performed by the subject, with the sEMG sensors placed on the biceps and triceps muscle groups. The processed sEMG signals were transformed as described in Section 2.1 to obtain muscle activation signals, used to calculate the MSK forward dynamics ODE residual. The same simplified rigid body model was used as in Section 4 and appropriately scaled anthropometric properties (for the geometry of the model) and physiological parameters (for muscle-tendon material models used for the muscle groups) based on the generic upper body model defined in [68,69] were used. Fig. 9 shows the measured data of the three trials, including the transient raw sEMG signals and the corresponding angular motion of the elbow flexion-extension of the subject.

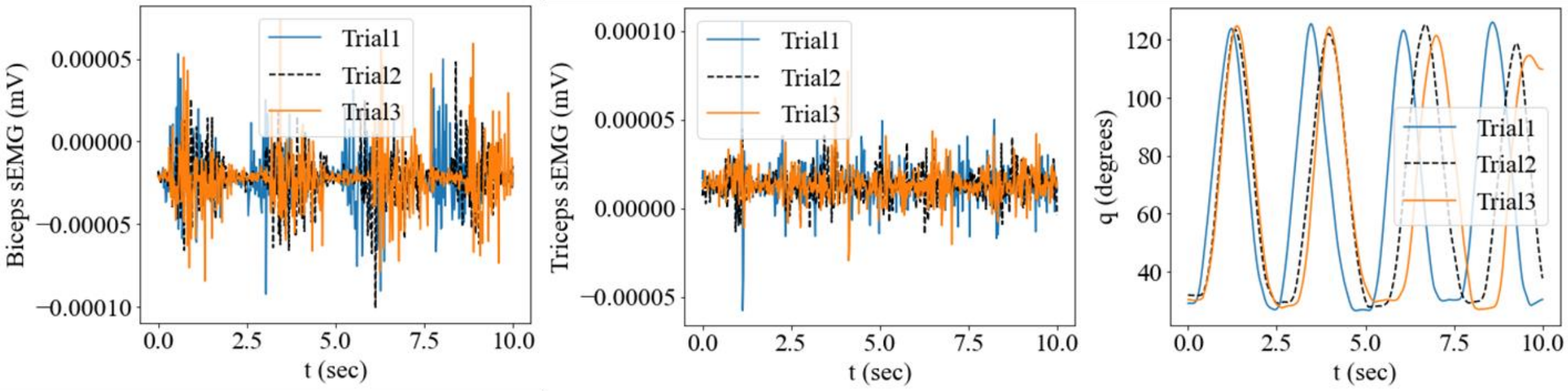

**Fig. 9**: The measured raw sEMG signals and the corresponding angular motion of the elbow flexion-extension of the subject are plotted.

In this example, the raw sEMG signals were used as input. A 5-scale MR training procedure as described in Section 4 was used on a GRU with 1 hidden layer with 50 neurons. The data of trials 1 and 3 were used for training, while trial 2 was used for testing, where each signal contained 500 temporal data points.

### 5.1.1 Parameter Identification

The muscle parameters to be identified by the framework include the maximum isometric force and the optimal muscle length from both muscle groups, which are denoted as $\boldsymbol{\Gamma} = \{f_{0,Bi}^M, l_{0,Bi}^M, f_{0,Tri}^M, l_{0,Tri}^M\}$. It was observed in our tests that despite the normalization process described in Eq. (37) and (38), the parameters obtained at the end of the MR training with motion data either diverged or converged to non-physiological values. To obtain physiologically consistent parameters, we use the values obtained from literature studies and constrain the space of parameter search [44].

Let the parameter to be identified be defined as

$$\Gamma_l(\boldsymbol{\psi}) = \frac{1}{N}\sum_{r=1}^{N} \bar{\gamma}_r \operatorname{sig}(\psi_r)\,, \boldsymbol{\psi} = [\psi_1, \psi_2, \dots, \psi_N] \tag{41}$$

where $\bar{\gamma}_r$ is the value defined in the $r^{th}$ literature study and $\psi_r$ is the parameter to be optimized in the training such that it can be used to evaluate the sigmoid function $\operatorname{sig}(\psi_r)$ and $\boldsymbol{\psi}$ is the vector of these trainable parameters. Using the optimized $\boldsymbol{\psi}$, the desired MSK parameters can be estimated. This formulation constrains the identified parameters to be consistent with parameters obtained through experimental studies [68–70].

The proposed framework was then applied to simultaneously learn the MSK forward dynamics surrogate and identify the MSK parameters $\boldsymbol{\Gamma}$ by optimizing Eq. (16), where the residual of the governing equation for the current time step $k$, was expressed as

$$\begin{aligned} & r\left(\hat{q}_k^{[-j]}(\boldsymbol{\theta_q}), \dot{\hat{q}}_k^{[-j]}(\boldsymbol{\theta_q}), \ddot{\hat{q}}_k^{[-j]}(\boldsymbol{\theta_q}); \boldsymbol{\Gamma}(\boldsymbol{\psi})\right) \\ & = I\ddot{\hat{q}}_k^{[-j]}(\boldsymbol{\theta_q}) - E\left(\hat{q}_k^{[-j]}(\boldsymbol{\theta_q})\right) - T^{MT}\left(a_{Bi}(t_k), a_{Tri}(t_k), \hat{q}_k^{[-j]}(\boldsymbol{\theta_q}), \dot{\hat{q}}_k^{[-j]}(\boldsymbol{\theta_q}); \boldsymbol{\Gamma}(\boldsymbol{\psi})\right) \end{aligned} \tag{42}$$

and could be plugged into the residual term $J_{res}$ in the loss function such that the optimization problem becomes,

$$\widetilde{\boldsymbol{\theta}}, \widetilde{\boldsymbol{\psi}} = \underset{\boldsymbol{\theta},\boldsymbol{\psi}}{\operatorname{argmin}}\big(J_{data}(\boldsymbol{\theta}) + \beta J_{res}(\boldsymbol{\theta}, \boldsymbol{\psi})\big). \tag{43}$$

As mentioned in the verification example (Section 4), the final parameter identification happens at the full-scale, i.e., at [0].

### 5.2 Results

The training was performed using the Adam algorithm [66] with an initial learning rate of $1 \times 10^{-3}$ and 4 history steps were considered. Multiple parameter initialization seeds were used for an averaged response of the MR training. To quantify the error in the testing predictions, a normalized mean squared error was defined,

$$\text{NMSE} = \frac{1}{n}\frac{\sum_{i=1}^{n}(q_i - \hat{q}_i)^2}{\sum_{i=1}^{n}(q_i - \bar{q})^2} \tag{44}$$

where $q_i$ is the $i^{th}$ target motion data point, $\hat{q}_i$ is the $i^{th}$ predicted motion data point, from the MR PI-RNN framework, and $\bar{q}$ is the mean of target motion data. The $R^2$ score was calculated using the metric defined in Eq. (40). From Fig. 10, Fig. 11 and Table 4, it is clear that addition of training scales leads to improved motion predictions. The multi-resolution training leads to an increase in average test $R^2$ score of more than 40% (bringing it closer to one), averaged over multiple initialization seeds. With the addition of more scales, Fig. 10 clearly shows the progression in improvement of the predictions as more scales are involved in the training.

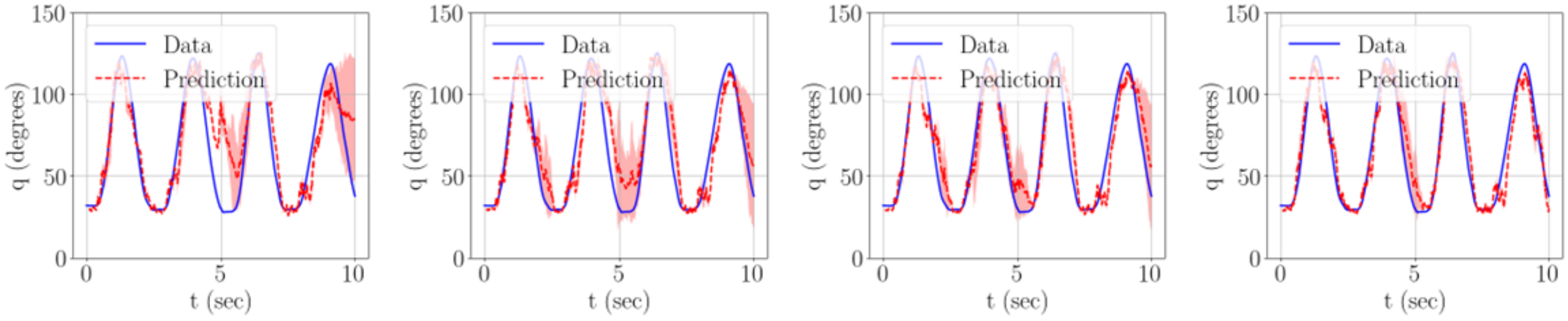

1-scale 2-scale 3-scale 4-scale

**Fig. 10**: Comparison of test predictions post-training for each MR training scale performed. The solid dash line is the mean of the predictions post-training when various initialization points are utilized to begin the MR training, with shaded region indicating one standard deviation from the mean.

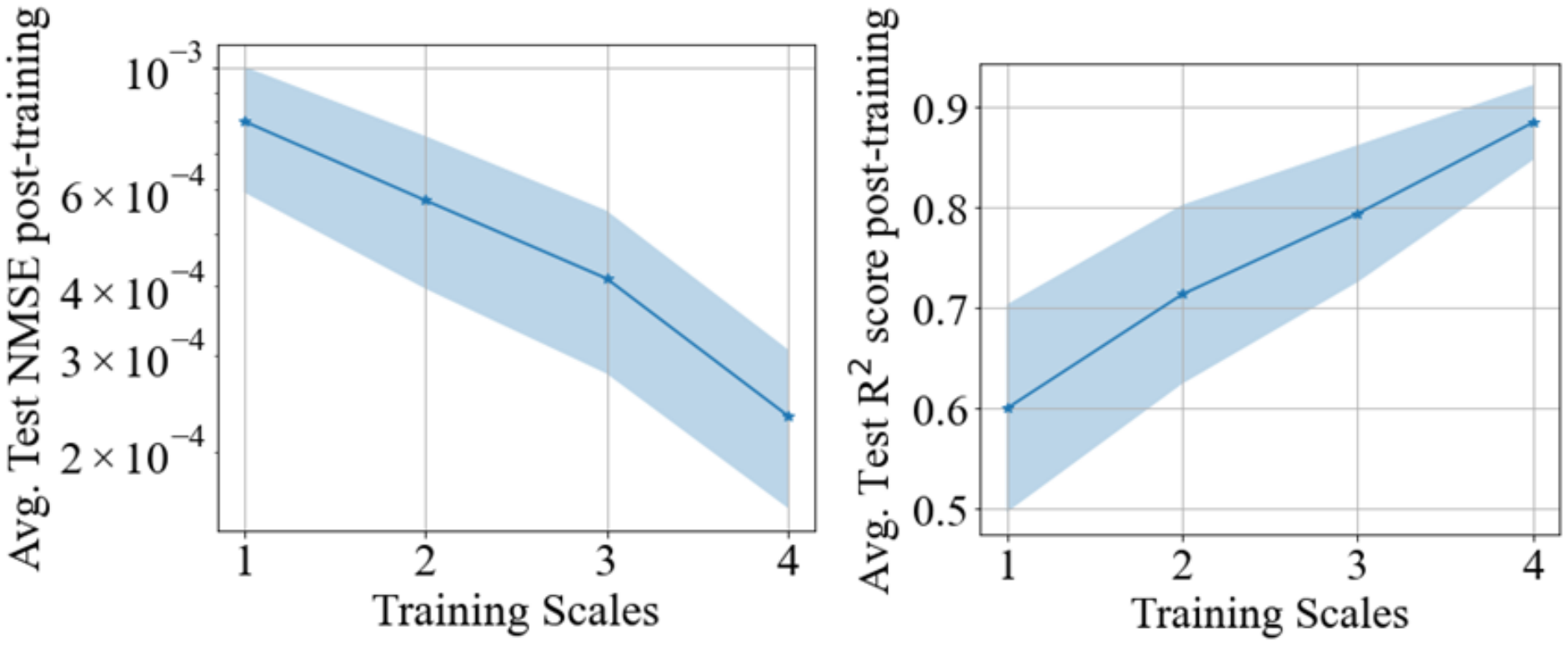

**Fig. 11**: The test normalized mean squared error (NMSE) and test $R^2$ score are plot for the testing predictions post-training, averaged over multiple initialization seeds. The mean of the metric is the solid marker line, the shaded portion being one standard deviation from the mean.

Table 4: The test metrics such as NMSE and $R^2$ score averaged over multiple initialization seeds, for the various training scales involved are reported here. The % decrease in average NMSE and % increase in average $R^2$ w.r.t 1-scale training are shown in the 3rd and 5th columns respectively.

| **Training Scale** | **Avg. Test NMSE** | **Decrease (%) in Avg. Test NMSE w.r.t 1-scale Training** | **Avg. Test $R^2$ Score** | **Increase (%) in Avg. Test $R^2$ score w.r.t 1-Scale training** |
|---|---|---|---|---|
| 1 | 8.00E-04 | - | 0.599 | - |
| 2 | 5.74E-04 | 28% | 0.713 | 19% |
| 3 | 4.14E-04 | 48% | 0.793 | 32% |
| 4 | 2.33E-04 | 71% | 0.884 | 47% |

The identified MSK parameters from the MR PI-RNN training are summarized in Table 5 with the mean of the final converged values of $f_0^M$ and $l_0^M$ obtained from multiple parameter initializations at 4-scale training, consistent with the physiological estimates of these parameters reported in literature [68–70]. $l_{0,Bi}^M$ is slightly outside the estimated range, which could be attributed to the variance in population. Similar values were obtained across all scales of training hence parameters obtained from a representative 4-scale training are shown here. The results demonstrated the effectiveness of the proposed MR PI-RNN framework and promising potential for real applications.

Table 5: The identified parameter estimates using MR PI-RNN training, and their values reported in literature [68–70].

| **Parameter** | **Identified values** | **Estimates from literature** |
|---|---|---|
| $f_{0,Bi}^M$ (N) | $348.23 \pm 0.2$ | 158.4-845 |
| $l_{0,Bi}^M$ (m) | $0.108 \pm 0.001$ | 0.115-0.142 |
| $f_{0,Tri}^M$ (N) | $758.36 \pm 0.5$ | 554.4-2332.916 |
| $l_{0,Tri}^M$ (m) | $0.069 \pm 0.001$ | 0.067-0.087 |

## 6. Discussion and Conclusions

In this work, we proposed a multi-resolution physics-informed recurrent neural network (MR PI-RNN) for an application to MSK systems, for time-domain motion prediction and parameter identification. A GRU with a physics-informed loss function that minimized the error in the training data and the residual of the MSK forward dynamics equilibrium was used for this purpose. Wavelet based multi-resolution techniques were used to decompose the input sEMG signals and output joint motion data into coarse-scale approximations at different scales and fine-scale details at those scales. The sEMG and joint motion multi-scale components were then mapped to each other starting from a chosen coarse-scale components and then sequentially trained (via transfer learning) to higher scales, completing the training on the full-scale of the data.

By initializing training on the coarse-scale of the training data, the optimization reaches a local minimum that serves as a better initialization state for the training data that includes the sequential fine-scale details. The proposed transfer-learning based sequential training scheme can be used for learning datasets that have high frequency signals as shown in the verification example with synthetic mixed frequency sEMG data. The numerical examples show an improvement in testing prediction and identifying the parameters. We observe from the loss profiles that the testing loss decreased while the training loss increased as more scales of data were brought in. It was also observed that the average test MSE and $R^2$ metrics showed a clear improvement in the generalization accuracy. These phenomena can be explained through the theory of bias-variance tradeoff; training on various scales of the data introduces more variance to the training, helping the ML framework to reduce the bias it develops by just training on the full-scale of the data. Computationally, it is noted that the proposed method achieves improved accuracy by using the same amount of training epochs.

The proposed MR framework was validated on recorded sEMG and motion data from a subject [1] and significant improvements were observed in the testing prediction accuracy, with 1-scale training often leading to large errors. The predicted motion at higher training scales showed improvements across all initialization points used, indicating the robustness of the method. The identified parameters were also consistent with the physiological range observed in literature.

This method also has the advantage of operating in the time-domain as compared to the feature-encoded (FE) training [1], where the input sEMG signals were projected on to the frequency domain using the Fourier basis. In the FE training, to make a prediction, the input signal for the entire duration of the movement prediction was needed whereas the physics informed MR training of the RNN enables the trained model to make real-time predictions by using the information of the previous time-steps and the current sEMG signal. In addition, for mixed frequency signals, wavelet resolution can better capture the local frequency information as compared to the Fourier basis which captures the global frequency information.

This method is presented as a general case where multi-resolution is applied to both input and output. For some applications, for e.g., those that require only data mapping, the MR training can be applied by only considering the decomposition to the input, keeping the output at the full-scale (i.e., scale [0]) throughout, or vice versa. To further improve this method, we can consider the use of multi-resolution as activation functions of the ML framework, instead of relying on data filtration processes for better computational efficiency. This method will also be tested on other physics-informed ML techniques to solve forward problems with PDEs having mixed frequency source terms.

## 7. Acknowledgment

The support of this work by the National Institute of Health under grant number 5R01AG056999-04 to K. Taneja & J. S. Chen are very much appreciated.

## Appendix A: Muscle-Tendon Force Generation

The total muscle force $F^M$ can be expressed as

$$F^M\left(a, \tilde{l}^M, \tilde{v}^M; \boldsymbol{\kappa}\right) = f_0^M \left(f^A\left(a, \tilde{l}^M, \tilde{v}^M; \boldsymbol{\kappa}\right) + f^P\left(\tilde{l}^M; \boldsymbol{\kappa}\right)\right), \tag{45}$$

where $f^P(\tilde{l}^M)$ is the passive muscle length dependent force generation function. The active force $f^A$ component can be expressed as:

$$f^A(a, \tilde{l}^M, \tilde{v}^M; \boldsymbol{\kappa}) = a f^{A,L}(\tilde{l}^M; \boldsymbol{\kappa}) f^V(\tilde{v}^M; \boldsymbol{\kappa}), \quad (46)$$
$$\tilde{l}^M = l^M / l_0^M,$$
$$\tilde{v}^M = v^M / v_{max}^M,$$

where $a$ is the activation function in Eq. (2), $\tilde{l}^M$ is the normalized muscle length, $\tilde{v}^M$ is the normalized velocity of the muscle. The total length of the MT system $l^{MT}$ is given by,

$$l^{MT} = l^M \cos\phi + l^T. \quad (47)$$

Given the current joint angle $q$ and the angular velocity $\dot{q}$, the current length, $l^{MT}$ of the MT system can be calculated using trigonometric relations.

The $f^{A,L}(\tilde{l}^M)$ and $f^V(\tilde{v}^M)$ are generic functions of the length and velocity dependent force generation properties of the active muscle, represented by dimensionless quantities. In this study, the tendon is assumed to be rigid ($l^T = l_s^T$). The total force produced by the MT complex, $F^{MT}$, can be expressed as:

$$F^{MT}(a, \tilde{l}^M, \tilde{v}^M, \phi; \boldsymbol{\kappa}) = F^M(a, \tilde{l}^M, \tilde{v}^M; \boldsymbol{\kappa}) \cos\phi. \quad (48)$$

The rigid-tendon model simplifies the MT contraction dynamics [57,58] which accounts for the interaction of the activation, force length, and force velocity properties of the MT complex.

**Appendix B. Hill-Type Muscle Models**

For the length dependent muscle force relations, this work uses the equations given in [56]. The active muscle force dependent on variation in length is given as

$$f^{A,L}(\tilde{l}^M) = \begin{cases} 9(\tilde{l}^M - 0.4)^2, & \tilde{l}^M \leq 0.6 \\ 1 - 4(1 - \tilde{l}^M)^2, & 0.6 \leq \tilde{l}^M \leq 1.4 \\ 9(\tilde{l}^M - 1.6)^2, & \tilde{l}^M > 1.4 \end{cases} \quad (49)$$

$$f^P(\tilde{l}^M) \quad (50)$$

$$= \begin{cases} 0, & \tilde{l}^M \leq 1 \\ \gamma_1\big(\exp\big(\gamma_2(\tilde{l}^M - 1)\big) - 1\big), & 1 \leq \tilde{l}^M \leq 1.4 \\ (\gamma_1\gamma_2 \exp(0.4\gamma_2))\tilde{l}^M + \gamma_1\big((1 - 1.4\gamma_2)\exp(0.4\gamma_2) - 1\big), & \tilde{l}^M > 1.4 \end{cases}$$

Where $\gamma_1 = 0.075$ and $\gamma_2 = 6.6$ correspond to parameters in the passive muscle force model related to an adult human. The muscle force velocity relationship $f^V(\tilde{v}^M)$ is used directly from [67].

## Appendix C: Multi-Resolution GRU Formulation

For multi-resolution GRU, the initial learning starts from the coarsest scale $[-j]$ as follows with notations according to Section 3.3.2,

$$\begin{aligned} \boldsymbol{r}_i^{[-j]} &= a_\sigma\left(\boldsymbol{W}_{hr}^{[-j]}\boldsymbol{h}_{i-1}^{[-j]} + \boldsymbol{W}_{xr}^{[-j]}\boldsymbol{x}_i^{[-j]} + \boldsymbol{W}_{qr}^{[-j]}\boldsymbol{q}_i^{[-j]} + \boldsymbol{b}_r^{[-j]}\right) \\ \boldsymbol{u}_i^{[-j]} &= a_\sigma\left(\boldsymbol{W}_{hu}^{[-j]}\boldsymbol{h}_{i-1}^{[-j]} + \boldsymbol{W}_{xu}^{[-j]}\boldsymbol{x}_i^{[-j]} + \boldsymbol{W}_{qu}^{[-j]}\boldsymbol{q}_i^{[-j]} + \boldsymbol{b}_u^{[-j]}\right) \\ \widetilde{\boldsymbol{h}}_i^{[-j]} &= a_{tanh}\left(\boldsymbol{r}_i^{[-j]} \odot \boldsymbol{W}_{h\tilde{h}}^{[-j]}\boldsymbol{h}_{i-1}^{[-j]} + \boldsymbol{W}_{x\tilde{h}}^{[-j]}\boldsymbol{x}_i^{[-j]} + \boldsymbol{W}_{q\tilde{h}}^{[-j]}\boldsymbol{q}_i^{[-j]} + \boldsymbol{b}_{\tilde{h}}^{[-j]}\right) \\ \boldsymbol{h}_i^{[-j]} &= \boldsymbol{u}_i^{[-j]} \odot \boldsymbol{h}_{i-1}^{[-j]} + \left(\boldsymbol{1} - \boldsymbol{u}_i^{[-j]}\right) \odot \widetilde{\boldsymbol{h}}_i^{[-j]} + \boldsymbol{b}_h^{[-j]} \end{aligned} \quad (51)$$

$$\forall\, i = n - m, \dots, n - 1,$$

$$\begin{aligned} \boldsymbol{r}_n^{[-j]} &= a_\sigma\left(\boldsymbol{W}_{hr}^{[-j]}\boldsymbol{h}_{n-1}^{[-j]} + \boldsymbol{W}_{xr}^{[-j]}\boldsymbol{x}_n^{[-j]} + \boldsymbol{b}_r^{[-j]}\right) \\ \boldsymbol{u}_n^{[-j]} &= a_\sigma\left(\boldsymbol{W}_{hu}^{[-j]}\boldsymbol{h}_{n-1}^{[-j]} + \boldsymbol{W}_{xu}^{[-j]}\boldsymbol{x}_n^{[-j]} + \boldsymbol{b}_u^{[-j]}\right) \\ \widetilde{\boldsymbol{h}}_n^{[-j]} &= a_{tanh}\left(\boldsymbol{r}_n^{[-j]} \odot \boldsymbol{W}_{h\tilde{h}}^{[-j]}\boldsymbol{h}_{n-1}^{[-j]} + \boldsymbol{W}_{x\tilde{h}}^{[-j]}\boldsymbol{x}_n^{[-j]} + \boldsymbol{b}_{\tilde{h}}^{[-j]}\right) \\ \boldsymbol{h}_n^{[-j]} &= \boldsymbol{u}_n^{[-j]} \odot \boldsymbol{h}_{n-1}^{[-j]} + \left(\boldsymbol{1} - \boldsymbol{u}_n^{[-j]}\right) \odot \widetilde{\boldsymbol{h}}_n^{[-j]} + \boldsymbol{b}_h^{[-j]} \end{aligned} \quad (52)$$

$$\widehat{\boldsymbol{q}}_n^{[-j]} = \boldsymbol{W}_{h\hat{q}}^{[-j]}\boldsymbol{h}_n^{[-j]} + \boldsymbol{b}_q^{[-j]} \quad (53)$$

where the weights $\boldsymbol{W}_{hr}^{[-j]}, \boldsymbol{W}_{xr}^{[-j]}, \boldsymbol{W}_{qr}^{[-j]}, \boldsymbol{W}_{hu}^{[-j]}, \boldsymbol{W}_{xu}^{[-j]}, \boldsymbol{W}_{qu}^{[-j]}, \boldsymbol{W}_{h\tilde{h}}^{[-j]}, \boldsymbol{W}_{x\tilde{h}}^{[-j]}, \boldsymbol{W}_{q\tilde{h}}^{[-j]}$ and biases $\boldsymbol{b}_r^{[-j]}, \boldsymbol{b}_u^{[-j]}, \boldsymbol{b}_{\tilde{h}}^{[-j]}, \boldsymbol{b}_h^{[-j]}, \boldsymbol{b}_q^{[-j]}$ are trainable parameters.

## Appendix D: Equation of Motion of the Simplified MSK Model

The equation of motion for the rigid body system used in Section 4 is,

$$I\ddot{q} = E(q) + T^{MT}(a_{Bi}, a_{Tri}, q, \dot{q}; \boldsymbol{\kappa}_{Bi}, \boldsymbol{\kappa}_{Tri}) \tag{54}$$

where,

$$I = m_{fa} l_{fa}^2 \tag{55}$$

$$E(q) = -m_{fa} g l_{fa} \sin(q) \tag{56}$$

$$T^{MT}(a_{Bi}, a_{Tri}, q, \dot{q}; \boldsymbol{\kappa}_{Bi}, \boldsymbol{\kappa}_{Tri}) = T_{Bi}^{MT}(a_{Bi}, q, \dot{q}; \boldsymbol{\kappa}_{Bi}) - T_{Tri}^{MT}(a_{Tri}, q, \dot{q}; \boldsymbol{\kappa}_{Tri}) \tag{57}$$

$$T_{Bi}^{MT}(a_{Bi}, q, \dot{q}; \boldsymbol{\kappa}_{Bi}) = \frac{F_{Bi}^{MT}\left(a_{Bi}, \tilde{l}_{Bi}^{M}, \tilde{v}_{Bi}^{M}, \tilde{l}_{Bi}^{T}; \boldsymbol{\kappa}_{Bi}\right) l_{2,Bi} \sin(q)\, l_{1,Bi}}{l_{Bi}^{MT}(q)} \tag{58}$$

$$T_{Tri}^{MT}(a_{Tri}, q, \dot{q}; \boldsymbol{\kappa}_{Tri}) = \frac{F_{Tri}^{MT}\left(a_{Tri}, \tilde{l}_{Tri}^{M}, \tilde{v}_{Tri}^{M}, \tilde{l}_{Tri}^{T}; \boldsymbol{\kappa}_{Tri}\right) l_{2,Tri} \sin(q)\, l_{1,Tri}}{l_{Tri}^{MT}(q)}. \tag{59}$$